\documentclass[times,twocolumn,final,oneside]{elsarticle}

\usepackage{graphicx}
\usepackage{medima}
\usepackage{framed,multirow}
\usepackage{array}

\usepackage{amssymb}
\usepackage{latexsym}
\usepackage{tikz}

\usepackage{xcolor}

\usepackage[normalem]{ulem}

\usepackage{hyperref}
\usepackage{dblfloatfix}
\usepackage{amsmath}
\usepackage{cases}
\usepackage{amsfonts,amssymb}
\usepackage{mathrsfs}
\usepackage{url}
\usepackage{enumerate}
\usepackage{algorithm}
\usepackage{algorithmic}
\usepackage{bm}
\usepackage{caption}
\usepackage{float}
\usepackage{txfonts}
\usepackage{booktabs}
\usepackage{threeparttable}
\usepackage{fancyhdr}
\usepackage{lettrine}
\usepackage{natbib}
\newcommand{\diagcell}[2]{%
  {\color{black}
  \tikz[baseline=(c.base)]{
    \node[inner sep=0pt, outer sep=0pt,
          minimum width=3.4em, minimum height=2.2em] (c) {};

    \draw[draw=black, line width=0.6pt, line cap=rect]
      (c.north west) -- (c.south east);

    \node[anchor=south west, text=black, inner sep=0.6pt]
      at ([xshift=0.6pt,yshift=0.6pt]c.south west) {#1};
    \node[anchor=north east, text=black, inner sep=0.6pt]
      at ([xshift=-0.6pt,yshift=-0.6pt]c.north east) {#2};
  }}%
}

\usepackage{silence}
\usepackage[final]{microtype}


\usepackage{etoolbox}


\setkeys{Gin}{draft=false}

\newif\ifdoubleblind
\doubleblindfalse

\definecolor{newcolor}{rgb}{.8,.349,.1}

\journal{Medical Image Analysis}

\begin{document}

% \verso{Y. Xiao \textit{et~al.}}

\begin{frontmatter}

\title{DDS-UDA: Dual-Domain Synergy for Unsupervised Domain Adaptation in Joint Segmentation of Optic Disc and Optic Cup}%

\author[1]{Yusong Xiao}
\author[1]{Yuxuan Wu}
\author[1]{Li Xiao\corref{cor1}}
\ead{xiaoli11@ustc.edu.cn}
\author[2]{Gang Qu}
\author[3]{Haiye Huo}
\author[2]{Yu-Ping Wang}
\address[1]{MoE Key Laboratory of Brain-Inspired Intelligence Perception and Cognition, University of Science and Technology of China, Hefei 230052, China}
\address[2]{Department of Biomedical Engineering, Tulane University, New Orleans, LA 70118, USA}
\address[3]{School of Mathematics and Computer Sciences, Nanchang University, Nanchang 330031, China}

\cortext[cor1]{Corresponding author:}

\begin{abstract}
\begin{sloppypar}
Convolutional neural networks (CNNs) have achieved exciting performance in joint segmentation of optic disc and optic cup on single\nobreakdash-institution datasets. However, their clinical translation is hindered by two major challenges: limited availability of large\nobreakdash-scale, high\nobreakdash-quality annotations and performance degradation caused by domain shift during deployment across heterogeneous imaging protocols and acquisition platforms. While unsupervised domain adaptation (UDA) provides a way to mitigate these limitations, most existing approaches do not address cross\nobreakdash-domain interference and intra\nobreakdash-domain generalization within a unified framework. In this paper, we present the Dual\nobreakdash-Domain Synergy UDA (DDS\nobreakdash-UDA), a novel UDA framework that comprises two key modules. First, a bi\nobreakdash-directional cross\nobreakdash-domain consistency regularization module is enforced to mitigate cross\nobreakdash-domain interference through feature-level semantic information exchange guided by a coarse\nobreakdash-to\nobreakdash-fine dynamic mask generator, suppressing noise propagation while preserving structural coherence. Second, a frequency\nobreakdash-driven intra\nobreakdash-domain pseudo label learning module is used to enhance intra\nobreakdash-domain generalization by synthesizing spectral amplitude\nobreakdash-mixed supervision signals, which ensures high\nobreakdash-fidelity feature alignment across domains. Implemented within a teacher\nobreakdash-student architecture, DDS\nobreakdash-UDA disentangles domain\nobreakdash-specific biases from domain\nobreakdash-invariant feature\nobreakdash-level representations, thereby achieving robust adaptation to heterogeneous imaging environments. We conduct a comprehensive evaluation of our proposed method on two multi\nobreakdash-domain fundus image datasets, demonstrating that it outperforms several existing UDA based methods and therefore providing an effective way for optic disc and optic cup segmentation.
\end{sloppypar}
\end{abstract}

\begin{keyword}
% \KWD 
\vspace{-1.5pt}\\ \vspace{-1.5pt}Cross\nobreakdash-domain\\ \vspace{-1.5pt}Domain shift\\ \vspace{-1.5pt}Joint segmentation of optic disc and optic cup\\ \vspace{-1.5pt}Unsupervised domain adaptation
\end{keyword}

\end{frontmatter}

% Guard against the class/template re-enabling twoside during the title/frontmatter flow.
\makeatletter
\@twosidefalse
\@mparswitchfalse
\makeatother

%\linenumbers

%% main text
\section{Introduction}\label{Sec1}
In fundus images, the optic disc (OD) and optic cup (OC) are critical anatomical structures that reflect optic nerve head morphology and are closely related to the diagnosis and monitoring of ocular diseases such as glaucoma \citep{moris2023assessing, sears2019progressive}. Accurate OD/OC segmentation provides the foundation for reliable morphological assessment and subsequent clinical interpretation. However, manual segmentation of OC and OD is usually time consuming and suffers from considerable inter\nobreakdash-expert variability. Deep learning has recently been largely explored to automatically segment these regions. In particular, supervised learning (SL)\nobreakdash-based segmentation models have achieved promising performance using large\nobreakdash-scale labeled datasets for training \citep{fu2018joint, DBLP:journals/ijon/KuiJLPHZ25}, where the training and test data are generally assumed to be independent and identically distributed \citep{DBLP:journals/mia/SongCWZHZYZ24}. However, this assumption does not always hold in practice, e.g., when the data collected from multiple sites are acquired with different scanning devices and/or acquisition protocols \citep{DBLP:journals/tbe/Guan022}. Specifically, there is domain shift from training data to test data, leading to significant reduction in segmentation accuracy and stability.
\par To address this issue, domain adaptation has been proposed to migrate the model trained on the source domain to the target domain with fine\nobreakdash-tuning. Ideally, the domain adaptation approaches can use sufficient labeled data from the target domain to fine\nobreakdash-tune the pre\nobreakdash-trained model in the source domain \citep{dou2018unsupervised, DBLP:conf/miccai/ChartsiasJDT17}, but labeled target domain data are often untenable in real\nobreakdash-world clinical scenarios \citep{DBLP:journals/corr/abs-2311-01702, DBLP:journals/pami/DongCSFD24}. More recently, unsupervised domain adaptation (UDA) has attracted much attention in medical image segmentation \citep{DBLP:journals/tmi/XianLTZZLLY23, DBLP:journals/tmi/ZhaoZXZGZ23, DBLP:journals/corr/abs-2308-01265, DBLP:journals/mia/SunDX22}, since it is a common practical scenario when historical annotated cohorts can be retained for iterative model updates and it does not rely on target domain labels. Note that we focus on the standard UDA setting where labeled source data remain accessible during adaptation, which differs from source\nobreakdash-free domain adaptation (SFDA) that assumes the source data are unavailable and only a pretrained source model is provided. The UDA approaches are to minimize the influence of domain shift by learning domain\nobreakdash-invariant features. For example, generative adversarial network (GAN) based UDA approaches \citep{DBLP:journals/mia/ZhengZDZYCSLQ24, DBLP:journals/tmi/XianLTZZLLY23} have been explored to perform image style transfer, where multiple classifiers and discriminators of different types were stacked to achieve high\nobreakdash-dimensional feature alignment, facilitating the learning of domain\nobreakdash-invariant features. However, these approaches are often susceptible to training instability and mode collapse due to the adversarial nature of GANs, leading to suboptimal feature representations \citep{DBLP:journals/mia/ZhengZDZYCSLQ24}. To address the limitation, attention based UDA approaches have been developed \citep{DBLP:journals/tmi/JiC24}, where self\nobreakdash-attention or cross\nobreakdash-attention mechanisms were leveraged to capture long\nobreakdash-range dependencies and align cross\nobreakdash-domain features. They excel at modeling complex relationships but at the cost of high computations and complex parameter \citep{DBLP:conf/wacv/ValindriaPRLARR18}. More recently, diffusion-based UDA methods have been explored as an alternative generative paradigm for bridging domain gaps \citep{lin2024stable}. By formulating cross\nobreakdash-domain translation as a progressive denoising process, diffusion models often provide more stable optimization and improved structure preservation compared with adversarial image translation. However, the iterative sampling procedure usually introduces substantial computational overhead and inference latency, and the performance can be sensitive to diffusion hyperparameters.
\par Inspired by the strategies in semi\nobreakdash-supervised learning, UDA approaches can be improved by incorporating two paradigms (i.e., consistency regularization and pseudo label learning) for domain shift \citep{DBLP:conf/iccv/JingZLS23, DBLP:journals/cbm/WangFPS25}. Consistency regularization generally enforces that the model produces consistent predictions for semantically equivalent content under either (i) intra\nobreakdash-domain perturbations on target images (e.g., strong/weak augmentation, noise injection) or (ii) cross\nobreakdash-domain transformations between source and target samples. However, it may fail to adequately preserve fine\nobreakdash-grained structural relationships and semantic boundary sensitivity that are critical for anatomical fidelity \citep{DBLP:journals/tmi/WuWGLCZVOZ23}. Pseudo label learning uses unlabeled data in the target domain to generate pseudo labels through an iterative self\nobreakdash-training process, which are optimized as supervised signals in subsequent training. However, it remains susceptible to error propagation from noisy pseudo labels, particularly during initial training phases characterized by low confidence. Recent studies \citep{DBLP:conf/cvpr/KimJSJNK22, DBLP:conf/iccvw/TangLHZL23} have advocated a unified framework by combining consistency regularization with pseudo label learning yet overlooking the intrinsic synergistic coupling between the two paradigms. Furthermore, picking style pseudo label selection methods \citep{DBLP:conf/iccvw/TangLHZL23} rely on the selection strategy, which may introduce excessive noise.
\par To address these limitations, we propose a novel UDA framework for medical image segmentation by combining intra\nobreakdash-domain pseudo label learning with bi\nobreakdash-directional cross\nobreakdash-domain consistency regularization based on a teacher\nobreakdash-student architecture. First, a frequency\nobreakdash-driven pseudo label learning pathway is employed to improve learning stability by reducing noise and inconsistencies in individual images. Additionally, we propose a dynamic mask generation mechanism to achieve the cross\nobreakdash-domain consistency regularization. The dynamic mask guides bi\nobreakdash-directional feature mixing between source and target domains, which enables integrating cross\nobreakdash-domain information while suppressing domain\nobreakdash-specific noise. By jointly optimizing these components, the proposed framework learns domain\nobreakdash-invariant feature\nobreakdash-level representations and achieves accurate segmentation without requiring target domain labels.
\par The contributions of this paper are summarized as follows.
\begin{itemize}
\item We propose a novel dual\nobreakdash-domain synergy UDA framework namely ``DDS\nobreakdash-UDA" to combine bi\nobreakdash-directional cross\nobreakdash-domain consistency regularization with intra\nobreakdash-domain pseudo label learning. This is achieved with a teacher\nobreakdash-student architecture to ensure robust adaptation to multi\nobreakdash-domain medical image segmentation.
\item We propose frequency\nobreakdash-driven intra\nobreakdash-domain pseudo label learning that augments intra\nobreakdash-domain diversity through spectral amplitude mixing, thereby providing high\nobreakdash-fidelity
supervision for cross\nobreakdash-domain feature alignment.
\item We design dynamic cross\nobreakdash-domain consistency regularization to suppress cross\nobreakdash-domain interference by bi\nobreakdash-directionally exchanging semantics. It contains a dynamic mask generator to progressively increase sensitivity to key anatomical structures from coarse to fine scales.
\item Our DDS\nobreakdash-UDA is comprehensively validated on two publicly available multi\nobreakdash-domain fundus image segmentation datasets, and demonstrated to significantly outperform state\nobreakdash-of\nobreakdash-the\nobreakdash-art methods.
\end{itemize}

\section{Related Work}\label{Relate_Work}

\subsection{Joint Segmentation of OC and OD}
In fundus images, OC and OD are critical anatomical features, which are associated with the diagnosis of conditions such as glaucoma \citep{moris2023assessing, sears2019progressive}. Therefore, it is indispensable to investigate the joint segmentation of OC and OD in fundus images, and there have been significant research efforts to develop more robust and accurate segmentation methods. Ding et al. \citep{DBLP:conf/miccai/DingYWDXCL20} and Bhattacharya et al. \citep{DBLP:journals/bspc/BhattacharyaHCPCD23} improved the attention module from different perspectives to enhance the model's respective field for extracting multi\nobreakdash-scale contextual information, thus capturing global contextual information. Guo et al. \citep{DBLP:journals/cbm/GuoLLTHC22} enhanced the representation of semantic information and suppressed irrelevant background features by combining channel attention mechanisms and spatial attention mechanisms. Li et al. \citep{DBLP:journals/cbm/LiZHH23} improved fundus and optic cup segmentation across varying scanners and resolutions using a transformer\nobreakdash-based backbone with auxiliary classifiers to refine details, though at the cost of increasing training complexity. In order to reduce parameter redundancy and improve feature coherence, Chen et al. \citep{DBLP:journals/titb/ChenPYWX24} suppressed the incoherent expression of feature information by adding a feature fusion module to U\nobreakdash-Net. Inspired by the strong performance of diffusion models in generative tasks, Lin et al. \citep{lin2024stable} proposed SDSeg, a segmentation framework built upon Stable Diffusion. It introduces a simple latent estimation strategy to enable a single-step reverse process and employs latent-fusion concatenation to eliminate the need for multiple samples. Despite their success in controlled settings, supervised approaches face substantial challenges in maintaining performance amid severe domain shifts \citep{DBLP:conf/miccai/HuLX22, DBLP:journals/eswa/DongFLPC22}. These limitations motivate the development of novel unsupervised domain adaptation (UDA) techniques.

\subsection{Unsupervised Domain Adaptation}
Domain shift is common in medical images due to different data distributions. Its existence has made domain adaptation a popular choice. Domain adaptation can be categorized into three types: supervised domain adaptation (SDA), semi\nobreakdash-supervised domain adaptation (SSDA), and UDA \citep{DBLP:journals/tbe/Guan022}. UDA currently dominates the field, given the paucity of labeled medical images. UDA strategies are roughly classified into image\nobreakdash-based alignment, feature\nobreakdash-based alignment, and hybrid\nobreakdash-based alignment \citep{DBLP:journals/corr/abs-2311-01702}. The core of image\nobreakdash-based alignment in UDA is to reduce the distribution difference between the target and source domains by adjusting the image styles, instead of using the matching data that preserve the contextual semantic information excellently. CycleGAN \citep{DBLP:conf/iccv/ZhuPIE17} implements cross\nobreakdash-domain transformations from source to target domain or from target to source domain without paired inputs and outputs. Consequently, there are growing studies to add other components on the base of CycleGAN to further minimize the domain discrepancy. Chartsias et al. \citep{DBLP:conf/miccai/ChartsiasJDT17} proposed a strategy based on the CycleGAN architecture to enable the generation of another modality from one image modality, such as CT synthesis from MRI. The multi\nobreakdash-organ segmentation network, SynSeg\nobreakdash-Net \citep{DBLP:journals/tmi/HuoXMBAMSAL19}, consisted of a CycleGAN\nobreakdash-based periodic synthesis network and a segmentation network, which not only generated high\nobreakdash-quality target domain data but also achieved end\nobreakdash-to\nobreakdash-end segmentation of target domain images. However, these image\nobreakdash-level methods can easily lead to structural deformation of the converted images with the difficulty in training. For feature\nobreakdash-based alignment in UDA, recent approaches attempted to minimize the feature differences between different domains or to efficiently extract the domain\nobreakdash-invariant features to achieve the alignment and transformation of cross\nobreakdash-domain data. Cui et al. \citep{DBLP:journals/cbm/CuiYJXZ21} employed a series of ways to gradually enhance the domain invariance between the extracted two domains using a shared encoder as well as incorporating a self\nobreakdash-attention mechanism into the GAN. Feature\nobreakdash-level alignment, however, ignores pixel\nobreakdash-level appearance shifts. Therefore, there is growing interest in hybrid\nobreakdash-based alignment by combining image\nobreakdash- and feature\nobreakdash-based alignments. Zhong et al. \citep{DBLP:journals/nca/ZhongZLLDS23} adopted an adaptive approach at both the image and the feature levels, which captures domain\nobreakdash-invariant features while transferring the source domain image style into the target domain image style, gradually minimizing the distribution difference between the target and source domains. Baldeon et al. \citep{baldeon2024stac} proposed an improved hybrid alignment adaptive framework. Specifically, CycleGAN was employed to perform image\nobreakdash-level alignment to force the original and converted images to be segmented in the same way, thus ensuring the consistency of semantic content in the image translation stage. Dual discriminators were then utilized to align the features of the original and transformed images in the semantic prediction space to generate more accurate semantic prediction. Existing hybrid\nobreakdash-based alignment methods \citep{DBLP:journals/nca/ZhongZLLDS23, baldeon2024stac} primarily operate in the spatial domain and rarely explore the complementary frequency‑domain property. In addition, relying on the predefined anatomical constraints of GANs makes it difficult to effectively extract cross\nobreakdash-domain features.

\subsection{Pseudo Label Learning and Consistency-Based Strategies}
Pseudo label learning and consistency regularization have gained attention in semi\nobreakdash-supervised learning (SSL). Classical works like Lee et al. \citep{lee2013pseudo} formalized pseudo labeling by assigning labels to unlabeled data via high\nobreakdash-confidence predictions. Tarvainen et al. \citep{DBLP:conf/nips/TarvainenV17} proposed consistency regularization through teacher\nobreakdash-student models that enforce prediction stability under perturbations (e.g., noise injection).

These principles have been extended to UDA to overcome domain shift. For pseudo label learning, Zheng et al. \citep{DBLP:journals/ijcv/ZhengY21} proposed a correction method based on prediction uncertainty, which dynamically adjusted the confidence threshold by modeling the prediction variance to suppress the interference of erroneous pseudo labels in model training. Zou et al. \citep{DBLP:conf/eccv/ZouYKW18} proposed a UDA framework based on iterative self\nobreakdash-training. This method alternates between generating pseudo labels on the target domain and retraining the segmentation model, which helps alleviate performance degradation caused by domain disparity. In parallel, consistency\nobreakdash-based UDA methods exploit domain\nobreakdash-shared invariances by enforcing agreement across different “views” of the data.  Wilhelm et al. \citep{DBLP:conf/wacv/TranhedenOPS21} proposed a UDA method based on cross\nobreakdash-domain hybrid sampling, which utilized a hybrid data enhancement strategy to strengthen cross\nobreakdash-domain feature consistency. Zhou et al. \citep{DBLP:conf/icmcs/0001ZYLM22} proposed the Regional Comparison Consistency Regularization (RCCR), which improved the robustness of the model to changes in complex environments. Specifically, it used a comparison learning strategy that drew the local area features of different images at the same location closer and pushed away the features of different areas. Independent pseudo label learning strategies tend to generate pseudo labels with insufficient confidence, whereas current consistency\nobreakdash-based UDA approaches predominantly employ static unidirectional mixing schemes which may worsen the overfitting risk.

\begin{figure*}[ht]
  \centering
  \includegraphics[width=\textwidth]{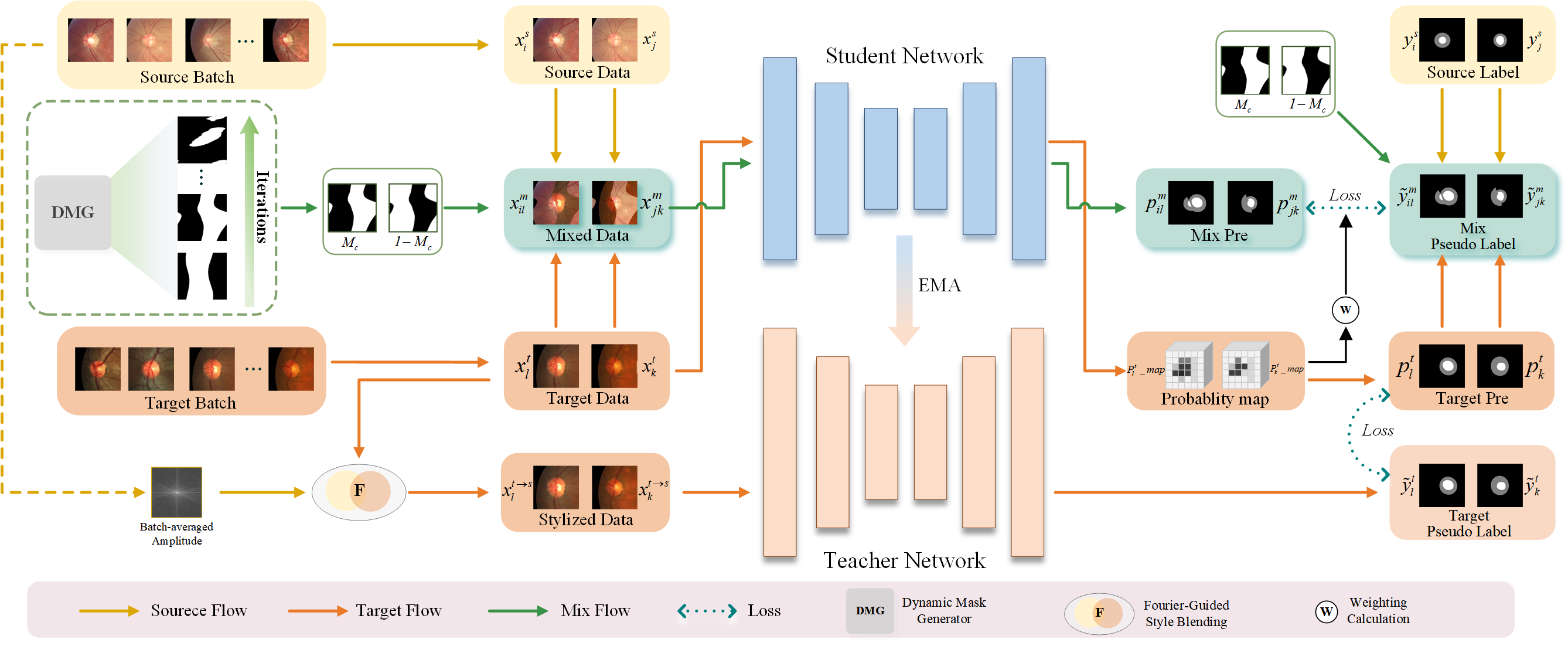}
  \caption{Overview of our proposed DDS\nobreakdash-UDA architecture, where we use the teacher\nobreakdash-student network as our basic network. Specifically, the bi\nobreakdash-directional cross\nobreakdash-domain consistency pathway (top) employs a dynamic mask generator (left) that facilitates bi\nobreakdash-directional feature-level semantic communication between the source and target domains by generating masks from coarse to fine, which are later trained by adding the same masks at loss computation in a consistency regularization manner. Meanwhile, the intra\nobreakdash-domain pseudo label learning pathway (bottom) extracts the average amplitude in the source domain data to synthesize stylized but semantically consistent target domain images, which are fed into the teacher model to generate pseudo labels to supervise the original target domain images.}
  \label{overview}
\end{figure*}

\section{Methods}\label{methods}

\subsection{Overview}\label{subsec3.1}

In the UDA setting, there are two domains, namely source domain $D_S$ and target domain $D_T$. Let $X_S = \{(x_i^s, y_i^s ),i = 1, ... , N_s\}$ be the training images and corresponding labels in the source domain, and $X_T = \{x_i^{t},i = 1, ... , N_t\}$ be the unlabeled images for adaptation in the target domain, where $N_s$ and $N_t$ are the numbers of samples in the two domains, respectively. Our goal is to mitigate the shift between $D_S$ and $D_T$ by UDA.

Fig. \ref{overview} illustrates our proposed DDS\nobreakdash-UDA, where we use a teacher\nobreakdash-student architecture and integrate a bi\nobreakdash-directional cross\nobreakdash-domain consistency regularization module and an intra\nobreakdash-domain pseudo\nobreakdash-label learning module to address domain shift. The intra\nobreakdash-domain pathway utilizes the Fourier transform to decompose an image into a phase spectrum and an amplitude spectrum. Stylistically migrated but semantically consistent enhancement samples are generated by batch averaging the source domain image amplitudes and mixing them with the target domain amplitudes. Then, pseudo labels generated from the stylized but semantically consistent images are used to supervise the original image, thereby providing higher\nobreakdash-fidelity supervisory signals for cross\nobreakdash-domain consistency regularization. The cross\nobreakdash-domain consistency regularization pathway employs a dynamic mask generator that produces the final mask to mix source and target domain data through Fourier\nobreakdash-based noise modulation, with a frequency attenuation factor in the process to adjust the mask granularity from coarse organ\nobreakdash-level regions to fine\nobreakdash-grained boundaries. These masks drive a bi\nobreakdash-directional copy\nobreakdash-paste processing that we copy masked source content into target images and vice versa, producing two hybrid images per pair, enabling symmetric feature exchange between source and target domains via fusing feature\nobreakdash-level structures. These two pathways interact in a unified framework, where cross\nobreakdash-domain hybrid features enhance the reliability of intra\nobreakdash-domain pseudo labels and stable pseudo labels guide cross\nobreakdash-domain alignment, forming an optimization loop that progressively adapts from global shape matching to local boundary refinement. This design stabilises early training under multi-centre shift by anchoring pseudo\nobreakdash-label learning and progressively controlling cross\nobreakdash-domain mixing, addressing a common issue in existing UDA frameworks where noisy early pseudo labels and overly aggressive mixing can reinforce errors. The implementation details of the pseudo label learning pathway and the consistency regularization pathway are provided in Sections \ref{subsec3.2} and \ref{subsec3.3}, respectively.

\subsection{Intra-domain Pseudo Label Learning}\label{subsec3.2}

Given an image $x$ of size $H\times W\times C$ (height, width, and the number of channels), its Fourier transform $F\triangleq\mathcal{F}(x)\in\mathbb{R}^{H\times W\times C}$ for the $c$-th channel is formulated as 

\begin{equation}\label{eq:FFT}F(u,v,c)= \sum\limits_{h = 0}^{H - 1} {\sum\limits_{w = 0}^{W - 1} {x} } (h,w,c){e^{ - {j2}\pi \left( {\frac{h}{H}u + \frac{w}{W}v} \right)}}.
\end{equation}

The Fourier spectrum $F$ can be decomposed as a product of an amplitude ${A} \in {\mathbb R^{H \times W \times C}}$ and a phase ${P} \in {\mathbb R ^{H \times W \times C}}$. The amplitude reflects low\nobreakdash-level characteristics (e.g., the image style), whereas the phase reflects high\nobreakdash-level semantics (e.g., the content of the image). Inspired by this property, we can modify the low-level appearance of an image by manipulating its amplitude while retaining its phase, which largely preserves semantic structure. This motivates a target\nobreakdash-to\nobreakdash-source style transfer for intra\nobreakdash-domain pseudo\nobreakdash-label learning. We merge the target amplitude with a batch-averaged source amplitude while keeping the target phase unchanged, so that the stylized target image better matches the source-domain appearance without altering the underlying OD/OC content. The teacher model predicts on this spectrum\nobreakdash-stylized view to obtain more stable pseudo labels, which are then used to supervise the student model on the original target image. 
\par First, let $\mathcal{S}=\{x_k^{s}\}_{k=1}^B\subset{X_S}$  and $\mathcal{T}=\{x_k^{t}\}_{k=1}^B\subset{X_T}$ be mini\nobreakdash-batches of source domain and target domain images, respectively, where $B$ denotes the batch size. By implementing the Fourier transform in (\ref{eq:FFT}) on every image of $\mathcal{S}$, we obtain the corresponding amplitudes $\left\{ {A}_k^s \right\}_{k = 1}^B$, based on which the average amplitude $A^s$ is given by
\begin{equation}
{ A^s} = \frac{1}{B}\sum\limits_{k = 1}^B {A_k^s}.
\end{equation}
For two randomly picked target domain images $\{x_l^t, x_k^t\} \subset \cal T$, we also obtain their amplitudes $\{A_l^t,A_k^t\}$ and phases $\{P_l^t, P_k^t\}$. We then introduce a zero\nobreakdash-centered binary mask $M$ to derive the spectrum\nobreakdash-stylized target domain images $x_{l}^{{{t}} \to {{s}}}$ and $x_{k}^{{{t}} \to {{s}}}$ by implementing the style transfer from the target domain to the source domain, i.e.,
\begin{align}\label{styletransfer}
{A}_l^{t \to s} &= {A}_l^t\odot{M} + {{{A}}^s}\odot({1- M}),\notag\\
{A}_k^{t \to s} &= {A}_k^t\odot{M} + {{{A}}^s}\odot({1- M}),\notag\\
x_{l}^{{{t}} \to {{s}}} &= {{\cal F}^{ - 1}}\left( {{A}_l^{t \to s}, {P}_l^t} \right),\notag\\
x_k^{{{t}} \to {{s}}}& = {{\cal F}^{ - 1}}\left( {{A}_k^{t \to s}, {P}_k^t} \right),
\end{align}
where $\odot$ and ${{\cal F}^{ - 1}}$ denote element\nobreakdash-wise multiplication and the inverse Fourier transform, respectively. Note that different from previous Fourier\nobreakdash-transform\nobreakdash-based style transfer approaches \citep{DBLP:journals/tmi/WuGWYYLZ24, DBLP:conf/eccv/ZhouQS22} to incorporate the amplitude of one random source domain image into the merged amplitude (namely ${A}_l^{t \to s}$ and ${A}_k^{t \to s}$) for each target domain image, our implementation in (\ref{styletransfer}) incorporates the average amplitude $A^s$  from all source domain images in $\mathcal{S}$, resulting in more robust spectrum\nobreakdash-stylized target domain images against style variations while alleviating the impact of noises and anomalies in a single source domain image.
\par Finally, since style transfer by only combining the amplitudes between the source and target domains does not affect the high\nobreakdash-level semantic information of images, the obtained spectrum\nobreakdash-stylized target domain images should keep the labels unchanged. To this end, we feed each target domain image and its spectrum\nobreakdash-stylized version into student and teacher networks, respectively, and enforce them to produce similar predictions for mitigating the influence of domain shift. Specifically, based on $x_l^{t}$ and $x_k^{t}$, we generate their probability maps $p_l^t\_\text{map}$ and  $p_k^t\_\text{map}$ from the student network, and subsequently take the argmax operation on them to acquire the corresponding predictions ${p}_l^{t}$ and ${p}_k^{t}$. Meanwhile, $x_l^{{{t}} \to {{s}}}$ and $x_k^{{{t}} \to {{s}}}$ are fed into the teacher network to generate pseudo labels $\tilde{{y}}_l^{t}$ and $\tilde{{y}}_k^{t}$, which provide supervisory signals to the student network.

\subsection{Cross\nobreakdash-domain Consistency Regularization}\label{subsec3.3}

We next present bi\nobreakdash-directional cross\nobreakdash-domain consistency regularization to minimize the domain discrepancy by combining the images from both the source and target domains in a structured manner below. For any two target domain images $\{x_l^t, x_k^t\}$ and two source domain images $\{x_i^s, x_j^s\}$, we leverage a binary mask $M_c$ to bi\nobreakdash-directionally blend the two pairs as follows.
\begin{align}\label{eq_M}
{x_{il}^{m} = x_i^s \odot  M_c + x_l^{t} \odot (1 -  M_c)},\notag\\
{x_{jk}^{m} =  x_j^s \odot (1 -  M_c)}+ x_k^{t} \odot   M_c,
\end{align}
where $x_{il}^{m}$ and $x_{jk}^{m}$ are referred to as mixed images, and $M_c$ is a customized mask of arbitrary shapes we will introduce later. From (\ref{eq_M}), we also have the pseudo labels of the mixed images 
\begin{align}\label{mixedlabel}
\tilde{{y}}_{il}^m&= {y}_i^s \odot {M_c} + {p}_l^{t} \odot ({1} - {M_c}),\notag \\
\tilde{{y}}_{jk}^m&=  {y}_j^s \odot ({1} - {M_c})+ {p}_k^{t} \odot {M_c},
\end{align}
where ${y}_i^s$ and ${y}_j^s$ denote the ground\nobreakdash-truth labels of $x_i^s$ and $x_j^s$ in the source domain, respectively. Similar to the operations on $x_l^{t}$ and $x_k^{t}$ to acquire ${p}_l^{t}$ and ${p}_k^{t}$ in Section \ref{subsec3.2}, we input $x_{il}^{m}$ and $x_{jk}^{m}$ into the student network to obtain their corresponding predictions ${p}_{il}^m$ and ${p}_{jk}^m$.

For the generation of the mask $M_c$ in (\ref{eq_M}) (without loss of generality, implemented in a single image channel configuration, i.e., $M_c\in\mathbb{R}^{H\times W}$), conventional mask generation approaches, e.g., CutMix \citep{cutmix}, are limited to only a rectangular region without the cross\nobreakdash-domain generalization. Following \citep{Fmix}, we develop a novel dynamic mask generation approach to derive a mask with variable shapes, which increases the ability to call for an intricate mixture of foreground and background. Specifically, let ${Z} \in \mathbb{C}^{H\times W}$ be a complex\nobreakdash-valued random matrix with its real and imaginary parts being independent and identically distributed as ${\Re(Z)} \sim \mathcal{N}(\mathbf{0}, \bm{I})$ and ${\Im(Z)} \sim \mathcal{N}(\mathbf{0}, \bm{I})$, respectively. Letting $\text{freq}(u,v)$ be the frequency magnitude at a 2D frequency grid point $(u,v)$, we compute it as the Euclidean norm of the vertical and horizontal frequency values that are uniformly distributed in the interval $\left[-\frac{1}{2}, \frac{1}{2}\right)$ on the two axes, respectively. We decay high\nobreakdash-frequency components in $Z$ via exponentially attenuating a coefficient proportional to the frequency magnitudes, i.e., 
\begin{equation}\label{ccc}
{Z_d}(u,v) = \frac{Z(u,v)}{\left(\max\left(\text{freq}(u,v), f_{\min}\right)\right)^d},
\end{equation}
where ${f_{\min }} = \frac{1}{{\max (H,W)}}$ is the minimum threshold to prevent division by zero, and $d$ stands for the frequency attenuation factor. Then, $M_c$ is generated by taking the inverse Fourier transform on $Z_d$, followed by the binarization with a threshold $\theta$, i.e.,
\begin{align}\label{ddd}
M_c&=\Re( {{\mathcal F}^{ - 1}}\left( Z_d \right)),\notag \\
{M_c(h,w)}& = \left\{ {\begin{array}{*{20}{l}}
{1,}&{{\rm{ if }}\;M_c(h,w) \ge \theta }\\
{0,}&{{\rm{ otherwise }}}
\end{array}} \right.,
\end{align}
where $\theta$ denotes the $k$\nobreakdash-th largest value in $ M_c$, and $k$ is defined as $k=\lambda_{k}HW$ with $\lambda_{k}$ being randomly generated within a preset interval.   From Eq. \ref{ccc} and Eq. \ref{ddd}, increasing $d$ will amplify the influence of low\nobreakdash-frequency components, producing a mask with smooth and large\nobreakdash-scale structures. Conversely, reducing $d$ will permit greater contributions from high\nobreakdash-frequency components, resulting in a mask with enhanced complexity and finer details. Thus, by controlling the value of $d$, we can adjust the frequency component distribution to generate a binary mask of different complexities and shapes. Here, we update $d$ during training by
\begin{equation}\label{di}
d_{i} = d_{\text{min}} + \left(1 - \frac{i}{I_{t}}\right) \cdot \left(d_{\text{max}} - d_{\text{min}}\right),
\end{equation}
where $d_{\text{max}}$, $d_{\text{min}}$, and $d_{i}$ respectively denote the maximum, minimum, and current (i.e., $i$\nobreakdash-th training iteration) values, and $I_t$ denotes the total number of training iterations. This strategy of gradually decreasing $d$ can give rise to irregularity in mask shapes, greatly improving the diversity of the mixed images to minimize the difference between the source and target domains.

\begin{table}[h]
\renewcommand{\arraystretch}{1.5}
\centering
\caption{A summary of the Fundus dataset used in this study.}
\resizebox{0.47\textwidth}{!}{
\begin{tabular}{l>{\large}c|>{\large}c>{\large}c>{\large}c>{\large}c}
\midrule
\multicolumn{1}{>{\large}c}{\multirow{2}{*}{\textbf{Domain}}} & & Domain 1 & Domain 2 & Domain 3 & Domain 4 \\
\cmidrule(lr){3-6}
& & Drishti-GS & RIM-ONE-r3 & REFUGE (train) & REFUGE (val) \\
\midrule
\multicolumn{1}{>{\large}c}{\textbf{Training}} & & 50 & 99 & 320 & 320 \\
\multicolumn{1}{>{\large}c}{\textbf{Testing}} & & 51 & 60 & 80 & 80 \\
\midrule
\multicolumn{1}{>{\large}c}{\textbf{Total}} & & 101 & 159 & 400 & 400 \\
\bottomrule
\end{tabular}
\label{tab_fundus}
}
\end{table}

\newcommand{\mstd}[2]{\ensuremath{#1_{\scriptscriptstyle #2}}}

\begin{table*}[b]
\centering
\scriptsize
\caption{Comparisons with advanced methods on the Fundus dataset. Metrics are reported as mean$_{\mathrm{std}}$ over three runs.}
\label{tab_Fundus}
\setlength{\tabcolsep}{2.5pt}
\renewcommand{\arraystretch}{1.5}
\resizebox{0.8\textwidth}{0.15\textheight}{
% ===================== Domains 1--2 =====================
\begin{tabular}{l|cccc|cccc}
\toprule
\multirow{2}{*}{Method} &
\multicolumn{4}{c|}{Domain 1} &
\multicolumn{4}{c}{Domain 2} \\
& $HD_{OD}$ $\downarrow$ & $Dice_{OD}$ $\uparrow$ & $HD_{OC}$ $\downarrow$ & $Dice_{OC}$ $\uparrow$
& $HD_{OD}$ $\downarrow$ & $Dice_{OD}$ $\uparrow$ & $HD_{OC}$ $\downarrow$ & $Dice_{OC}$ $\uparrow$ \\
\midrule

Source-only
& \mstd{9.53}{0.62} & \mstd{92.88}{0.41} & \mstd{14.01}{0.74} & \mstd{82.23}{0.53}
& \mstd{13.58}{0.69} & \mstd{87.54}{0.58} & \mstd{16.69}{0.83} & \mstd{78.69}{0.64} \\

Target-only
& \mstd{7.11}{0.45} & \mstd{97.15}{0.22} & \mstd{10.37}{0.56} & \mstd{87.77}{0.40}
& \mstd{8.71}{0.51} & \mstd{94.74}{0.33} & \mstd{9.77}{0.48} & \mstd{84.08}{0.52} \\

\midrule

FDA \cite{DBLP:conf/cvpr/0001S20}
& \mstd{9.78}{0.58} & \mstd{94.89}{0.36} & \mstd{14.39}{0.71} & \mstd{84.27}{0.55}
& \mstd{10.83}{0.60} & \mstd{89.79}{0.49} & \mstd{12.47}{0.63} & \mstd{81.11}{0.57} \\

DoCR \cite{DBLP:conf/miccai/HuLX22}
& \mstd{7.39}{0.47} & \mstd{95.87}{0.30} & \mstd{12.24}{0.66} & \mstd{85.33}{0.49}
& \mstd{9.98}{0.55} & \mstd{91.64}{0.44} & \mstd{11.59}{0.58} & \mstd{83.97}{0.50} \\

DAFormer \cite{hoyer2022daformer}
& \mstd{7.84}{0.61} & \mstd{95.12}{0.53} & \mstd{12.87}{0.66} & \mstd{84.95}{0.95}
& \mstd{9.96}{0.59} & \mstd{95.87}{0.80} & \mstd{12.01}{0.66} & \mstd{83.81}{0.87} \\

CoFo \cite{DBLP:conf/isbi/HuyHNDBT22}
& \mstd{8.42}{0.53} & \mstd{95.32}{0.33} & \mstd{13.29}{0.69} & \mstd{85.55}{0.51}
& \mstd{10.07}{0.59} & \mstd{90.38}{0.52} & \mstd{11.21}{0.60} & \mstd{83.39}{0.56} \\

BEAL \cite{DBLP:journals/cmpb/LiuPSS22}
& \mstd{7.62}{0.50} & \mstd{95.72}{0.32} & \mstd{12.91}{0.67} & \mstd{85.59}{0.50}
& \mstd{9.38}{0.57} & \mstd{89.13}{0.56} & \mstd{10.99}{0.61} & \mstd{82.48}{0.59} \\

ODADA \cite{DBLP:journals/mia/SunDX22}
& \mstd{8.87}{0.55} & \mstd{95.32}{0.34} & \mstd{12.36}{0.65} & \mstd{84.51}{0.54}
& \mstd{9.59}{0.58} & \mstd{90.62}{0.50} & \mstd{11.35}{0.62} & \mstd{81.72}{0.60} \\

TIST \cite{DBLP:conf/miccai/GhamsarianTMWZSS23}
& \mstd{9.29}{0.57} & \mstd{95.97}{0.31} & \mstd{13.66}{0.70} & \mstd{85.27}{0.52}
& \mstd{9.72}{0.60} & \mstd{89.32}{0.55} & \mstd{11.86}{0.63} & \mstd{82.38}{0.58} \\

MA\nobreakdash-UDA \cite{ji2023unsupervised}
& \mstd{8.48}{0.21} & \mstd{96.01}{0.27} & \mstd{12.71}{0.42} & \mstd{86.09}{0.82}
& \mstd{9.39}{0.39} & \mstd{90.21}{0.68} & \mstd{10.97}{0.55} & \mstd{82.69}{0.79} \\

TriLA \cite{DBLP:journals/titb/ChenPYWX24}
& \mstd{6.63}{0.40} & \mstd{96.21}{0.27} & \mstd{11.84}{0.57} & \mstd{86.69}{0.46}
& \mstd{7.63}{0.45} & \mstd{90.97}{0.48} & \mstd{9.97}{0.52} & \mstd{83.91}{0.55} \\

\midrule

\textbf{Ours}
& \textbf{\mstd{5.73}{0.21}} & \textbf{\mstd{96.62}{0.24}} & \textbf{\mstd{10.89}{0.21}} & \textbf{\mstd{87.46}{0.32}}
& \textbf{\mstd{8.54}{0.26}} & \textbf{\mstd{90.65}{0.36}} & \textbf{\mstd{10.77}{0.25}} & \textbf{\mstd{84.02}{0.23}} \\

\bottomrule
\end{tabular}%
}

\par\vspace{2mm}\par

% ===================== Domains 3--4 =====================
\resizebox{0.8\textwidth}{0.15\textheight}{
\begin{tabular}{l|cccc|cccc}
\toprule
\multirow{2}{*}{Method} &
\multicolumn{4}{c|}{Domain 3} &
\multicolumn{4}{c}{Domain 4} \\
& $HD_{OD}$ $\downarrow$ & $Dice_{OD}$ $\uparrow$ & $HD_{OC}$ $\downarrow$ & $Dice_{OC}$ $\uparrow$
& $HD_{OD}$ $\downarrow$ & $Dice_{OD}$ $\uparrow$ & $HD_{OC}$ $\downarrow$ & $Dice_{OC}$ $\uparrow$ \\
\midrule

Source-only
& \mstd{9.41}{0.61} & \mstd{90.78}{0.49} & \mstd{11.69}{0.67} & \mstd{82.19}{0.56}
& \mstd{11.74}{0.66} & \mstd{88.63}{0.55} & \mstd{15.81}{0.78} & \mstd{79.17}{0.63} \\

Target-only
& \mstd{5.13}{0.38} & \mstd{96.21}{0.26} & \mstd{5.96}{0.41} & \mstd{88.49}{0.39}
& \mstd{4.15}{0.34} & \mstd{95.72}{0.28} & \mstd{6.32}{0.43} & \mstd{90.10}{0.34} \\

\midrule

FDA \cite{DBLP:conf/cvpr/0001S20}
& \mstd{9.59}{0.60} & \mstd{91.41}{0.48} & \mstd{11.56}{0.66} & \mstd{84.49}{0.52}
& \mstd{9.42}{0.59} & \mstd{91.63}{0.50} & \mstd{11.39}{0.64} & \mstd{84.37}{0.54} \\

DoCR \cite{DBLP:conf/miccai/HuLX22}
& \mstd{9.42}{0.38} & \mstd{92.99}{0.43} & \mstd{10.77}{0.61} & \mstd{86.65}{0.47}
& \mstd{9.51}{0.58} & \mstd{93.18}{0.41} & \mstd{10.36}{0.60} & \mstd{86.03}{0.48} \\

DAFormer \cite{hoyer2022daformer}
& \mstd{9.31}{0.72} & \mstd{92.69}{0.57} & \mstd{10.67}{0.23} & \mstd{86.81}{0.58}
& \mstd{9.67}{0.69} & \mstd{93.76}{0.92} & \mstd{10.41}{0.56} & \mstd{86.34}{0.83} \\

CoFo \cite{DBLP:conf/isbi/HuyHNDBT22}
& \mstd{9.08}{0.56} & \mstd{93.08}{0.44} & \mstd{10.83}{0.63} & \mstd{86.98}{0.46}
& \mstd{6.85}{0.48} & \mstd{93.34}{0.43} & \mstd{8.99}{0.55} & \mstd{86.67}{0.47} \\

BEAL \cite{DBLP:journals/cmpb/LiuPSS22}
& \mstd{7.64}{0.51} & \mstd{92.08}{0.52} & \mstd{9.68}{0.60} & \mstd{86.34}{0.49}
& \mstd{8.38}{0.55} & \mstd{92.18}{0.50} & \mstd{9.87}{0.60} & \mstd{85.54}{0.52} \\

ODADA \cite{DBLP:journals/mia/SunDX22}
& \mstd{8.95}{0.57} & \mstd{92.14}{0.48} & \mstd{10.32}{0.62} & \mstd{85.78}{0.50}
& \mstd{9.76}{0.60} & \mstd{91.97}{0.49} & \mstd{11.54}{0.65} & \mstd{85.29}{0.52} \\

TIST \cite{DBLP:conf/miccai/GhamsarianTMWZSS23}
& \mstd{8.87}{0.56} & \mstd{92.67}{0.46} & \mstd{10.54}{0.63} & \mstd{86.22}{0.48}
& \mstd{8.96}{0.57} & \mstd{92.36}{0.47} & \mstd{10.05}{0.61} & \mstd{85.88}{0.49} \\

MA\nobreakdash-UDA \cite{ji2023unsupervised}
& \mstd{7.68}{0.66} & \mstd{93.01}{0.31} & \mstd{9.61}{0.42} & \mstd{86.86}{0.73}
& \mstd{8.21}{0.72} & \mstd{93.25}{0.48} & \mstd{9.37}{0.98} & \mstd{86.91}{0.86} \\

TriLA \cite{DBLP:journals/titb/ChenPYWX24}
& \mstd{6.45}{0.44} & \mstd{93.71}{0.45} & \mstd{8.73}{0.55} & \mstd{87.52}{0.43}
& \mstd{7.68}{0.50} & \mstd{93.78}{0.44} & \mstd{8.57}{0.54} & \mstd{86.69}{0.46} \\

\midrule

\textbf{Ours}
& \textbf{\mstd{6.30}{0.45}} & \textbf{\mstd{\textbf{94.45}}{0.40}} & \textbf{\mstd{8.89}{0.53}} & \textbf{\mstd{88.05}{0.41}}
& \textbf{\mstd{5.01}{0.39}} & \textbf{\mstd{94.24}{0.42}} & \textbf{\mstd{7.91}{0.50}} & \textbf{\mstd{87.40}{0.42}} \\

\bottomrule
\end{tabular}%
}
\end{table*}

\subsection{Loss Function}\label{subsec3.4}

For the spectrum\nobreakdash-stylized target domain images, the outputs $\tilde{{y}}_l^{t}$ and $\tilde{{y}}_k^{t}$ from the teacher network are leveraged to guide the student network to produce predictions ${p}_l^{t}$ and ${p}_k^{t}$ similar with them in the intra\nobreakdash-domain pseudo label learning. To this end, a hybrid loss function $\mathcal{L}_{\text{ipl}}$ consisting of the Intersection over Union (IoU) loss and Cross\nobreakdash-Entropy (CE) loss is proposed to minimize the difference between $\tilde{{y}}_l^{t}$ and ${p}_l^{t}$ as well as the difference between $\tilde{{y}}_k^{t}$ and ${p}_k^{t}$, i.e.,
\begin{equation}
\mathcal{L}_{\mathcal{T}\text{-stylized}}=\mathcal{L}_{\text{ipl}}\left({\tilde{{y}}_l^{t}}, {p}_l^t\right)+\mathcal{L}_{\text{ipl}}\left(\tilde{{y}}_k^{t}, {p}_k^t\right),
\end{equation}
where $\mathcal{L}_{\text{ipl}}$ combines the IoU loss and CE loss through a linear weighting scheme.

Likewise, for the mixed images in the bi\nobreakdash-directional cross\nobreakdash-domain consistency regularization, we encourage the outputs $\tilde{{y}}_{il}^m$ and $\tilde{{y}}_{jk}^m$ from the teacher network and the predictions ${p}_{il}^m$ and ${p}_{jk}^m$ from the student network to be similar. According to (\ref{mixedlabel}), we measure their differences from the source and target domains, respectively, using both the Dice and CE losses, i.e.,
\begin{equation}\label{e1}
\begin{aligned}
{{\cal L}_{{\mathcal{S}}}} = &{{\cal L}_{{{\text{Dice}}}}}\left( {y_i^s\odot {M_c},{{p}}_{{il}}^m}\odot {M_c} \right)  + {{\cal L}_{\text{CE}}}\left( {y_i^s\odot {M_c},{{p}}_{il}^m}\odot {M_c} \right) \\&+  {{\cal L}_{{{\text{Dice}}}}}\left( {y_j^s\odot {(1-M_c)}, {{p}}_{{jk}}^m}\odot {(1-M_c)} \right) \\& + {{\cal L}_{\text{CE}}}\left( {y_j^s\odot {(1-M_c)},{{p}}_{{jk}}^m}\odot {(1-M_c)} \right),  
\end{aligned}
\end{equation}
\begin{equation}\label{e2}
\begin{aligned}
{{\cal L}_{{\mathcal{T}}}} = &\gamma_l\cdot{{\cal L}_{{{\text{Dice}}}}}\left( {p_l^t\odot {(1-M_c)},{{p}}_{{il}}^m}\odot {(1-M_c)} \right) \\ &+ \gamma_l\cdot{{\cal L}_{\text{CE}}}\left( {p_l^t\odot {(1-M_c)},{{p}}_{{il}}^m}\odot {(1-M_c)} \right) \\&+ \gamma_k\cdot {{\cal L}_{{{\text{Dice}}}}}\left( {p_k^t\odot {M_c}, {{p}}_{{jk}}^m}\odot {M_c} \right) \\& + \gamma_k\cdot{{\cal L}_{\text{CE}}}\left( {p_k^t\odot {M_c}, {{p}}_{{jk}}^m}\odot {M_c} \right).
\end{aligned}
\end{equation}
As the ground truths $y_i^s$ and $y_j^s$ of the source domain images are believed to be more accurate than the predictions $p_l^t$ and $p_k^t$ of the target domain images obtained by the student network, we introduce weighting parameters $\gamma_l$ and $\gamma_k$ in (\ref{e2}) to weaken the impact of ${{\cal L}_{{\mathcal{T}}}}$ with low confidence, where
\begin{equation}
\gamma_l = \frac{1}{{H \times W}}\sum\limits_{h = 0}^{H-1} {\sum\limits_{w = 0}^{W-1} {\mathop {\max }\limits_{_{c \in \{ 1,2, \ldots ,C\} }} } } \{p_l^t\_\text{map}(h,w,c)\},
\end{equation}
$p_l^t\_\text{map}$ is the probability map from the student network, and $\gamma_k$ can be defined analogously. At each training iteration, we update the weights of the student network with the following total loss function
\begin{equation}
{\cal L}_{\text{total}} = {\lambda_{\mathcal{S}}\cdot{\cal L}_\mathcal{S}} + \lambda_{\mathcal{T}}\cdot {{\cal L}_\mathcal{T}} + \lambda_{\mathcal{T}\text{-stylized}} \cdot{\cal L}_{\mathcal{T}\text{-stylized}},
\end{equation}
where $\lambda_{\mathcal{S}}$, $\lambda_{\mathcal{T}}$, and $\lambda_{\mathcal{T}\text{-stylized}} $ are tunable parameters. The weights of the teacher network are updated based on the exponential moving average (EMA) algorithm, which is commonly applied in teacher\nobreakdash-student network architecture.

\begin{figure*}[t]
  \centering
  \includegraphics[width=\textwidth]{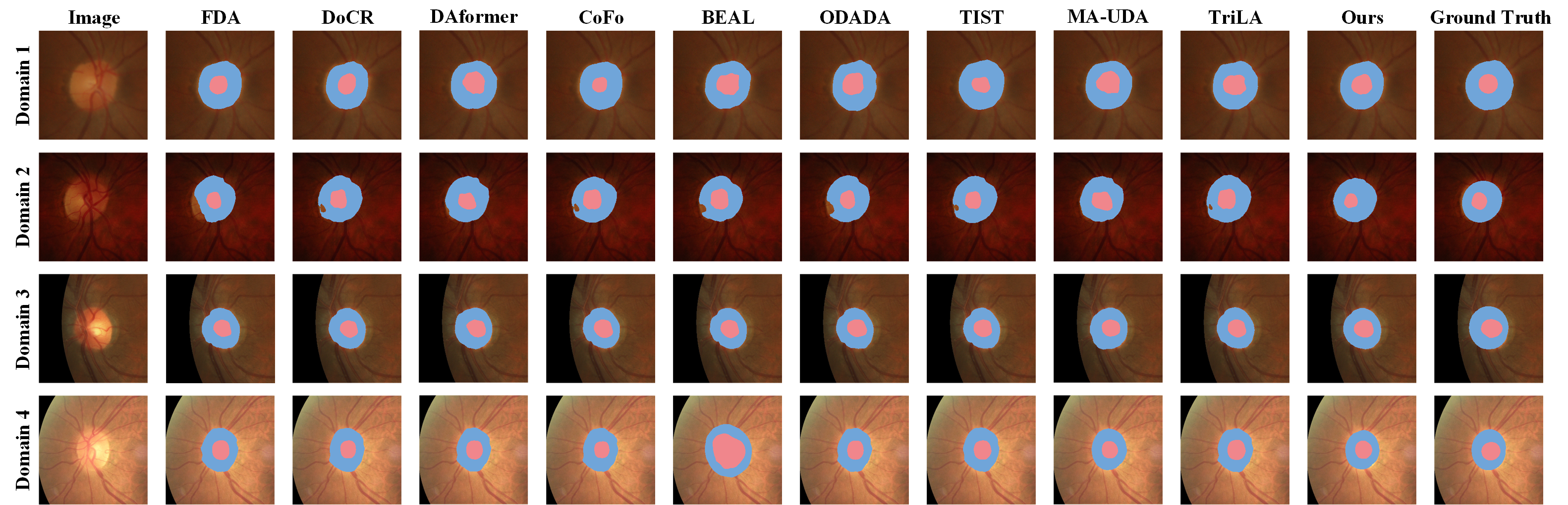}
  \caption{The visualization results produced by different methods on the Fundus dataset. Red and blue colors denote OC and OD, respectively.}
  \label{Fundus_visual}
\end{figure*}

\section{Experiments}\label{expem}
\subsection{Dataset and Implementation}

Following \cite{Wang2020}, we used the Fundus dataset, which is constructed from four public optic disc/cup segmentation datasets collected at different medical centres and with different cameras. Concretely, the four domains correspond to Drishti-GS, RIM-ONE-r3, and the training and validation subsets of the REFUGE challenge. Because of the differences in acquisition protocols, field of view, image resolution and illumination, these four datasets exhibited clear appearance discrepancies, and we therefore treated each of them as one domain. Each image was provided with expert pixel-wise annotations of the optic disc (OD) and optic cup (OC); these masks were used as the ground truth for all our experiments. The numbers of images for each domain and the corresponding train/test splits were summarized in \ref{tab_fundus}. As in \cite{Wang2020}, the images in Domains 1 and 2 followed the original training/testing partition provided by the datasets. For Domains 3 and 4, we randomly divided the images into training and testing subsets with a ratio of 4:1. In our unsupervised domain adaptation setting, we alternately chose one domain as the target domain and treated the remaining three domains as the multi-source domains. During adaptation, the model was trained on the labeled training subsets of the source domains together with the unlabeled training images of the target domain, and it was evaluated exclusively on the test subset of the target domain.
\par All the experiments were implemented with PyTorch, using an NVIDIA GeForce RTX 4070 GPU. For CNN-based methods, we selected the widely used ResNet \cite{he2016deep} with DeepLabV3+ \cite{chen2018encoder} to demonstrate the validity of our method. For transformer-based approaches, we kept the model architectures consistent with those in the original publications. In our experiments, the batch size was set to 8. For Eq. \ref{di}, we set $d_{\min}=1$ and $d_{\max}=5$. During pre\nobreakdash-training in the source domain, we trained the source model for 5000 iterations with Dice loss, the stochastic gradient descent (SGD) optimizer and a learning rate of 0.001. The model parameters with the best performance on the test set in the source domain were used for adaptation. At the next stage of domain adaptation, we trained the model for 12000 iterations and used poly scheduling to adjust the learning rate as $l = {l_{{\rm{init }}}} \cdot {\left( {1 - \frac{i}{I}} \right)^{0.9}}$, where $l_{{\rm{init }}}$ was the initial learning rate set to 0.001, $i$ was the current iteration, and $I$ was the maximum number of iterations.
\par Following \cite{Wang2020}, we center\nobreakdash-cropped a fixed 800$\times$800 window around the annotated disc center. After that, we randomly resized and cropped a 256$\times$256 region on each cropped image as network input. And we applied the basic data augmentation to expand the training samples, including random rotation and flipping. Moreover, we conducted the morphological operation, i.e., filling the hole, to post\nobreakdash-process the predicted masks. To reduce randomness, we repeated our experiments for three times and reported the average performance.

\subsection{Comparison with Competitive Methods}
To verify the effectiveness of our proposed DDS\nobreakdash-UDA, we compared it with 7 competitive UDA methods: 1) FDA \cite{DBLP:conf/cvpr/0001S20} that performed Fourier domain adaptation by transferring domain\nobreakdash-specific features in the frequency space for semantic segmentation; 2) DoCR \cite{DBLP:conf/miccai/HuLX22} that used domain\nobreakdash-specific contrastive learning to align source and target domain features for robust medical image segmentation; 3) DAFormer \cite{hoyer2022daformer} that improved domain-adaptive semantic segmentation by adopting a Transformer encoder together with a context-aware multi-level feature fusion decoder that stacks multi-level features and fuses them via parallel dilated depthwise separable convolutions; 4) CoFo \cite{DBLP:conf/isbi/HuyHNDBT22} that combined image translation and feature alignment for joint optimization in unsupervised domain adaptation; 5) BEAL \cite{DBLP:journals/cmpb/LiuPSS22} that employed bi\nobreakdash-directional feature alignment and adversarial learning for domain adaptation in medical imaging; 6) ODADA \cite{DBLP:journals/mia/SunDX22} that utilized disentangled representations to separate domain\nobreakdash-specific and share features for robust cross\nobreakdash-domain segmentation; 7) TIST \cite{DBLP:conf/miccai/GhamsarianTMWZSS23} that adopted task\nobreakdash-informed self\nobreakdash-training to enhance domain adaptation for medical image segmentation; 8) MA\nobreakdash-UDA \cite{ji2023unsupervised} that built a Transformer-based UDA framework for cross\nobreakdash-modality medical image segmentation by combining pixel-level alignment with attention-level alignment; and 9) TriLA \cite{DBLP:journals/titb/ChenPYWX24} that incorporated triplet loss and adversarial learning for improved domain adaptation in biomedical segmentation. We also compared these methods with two naive methods: 1)  ``Source\nobreakdash-only" which used only source domain data to train the network and test it directly on the target domain without domain adaptation; and 2) ``Target\nobreakdash-only" that only used training images and corresponding labels in the target domain to train and test the network.

\begin{figure*}[h]
  \centering
  \includegraphics[width=\textwidth]{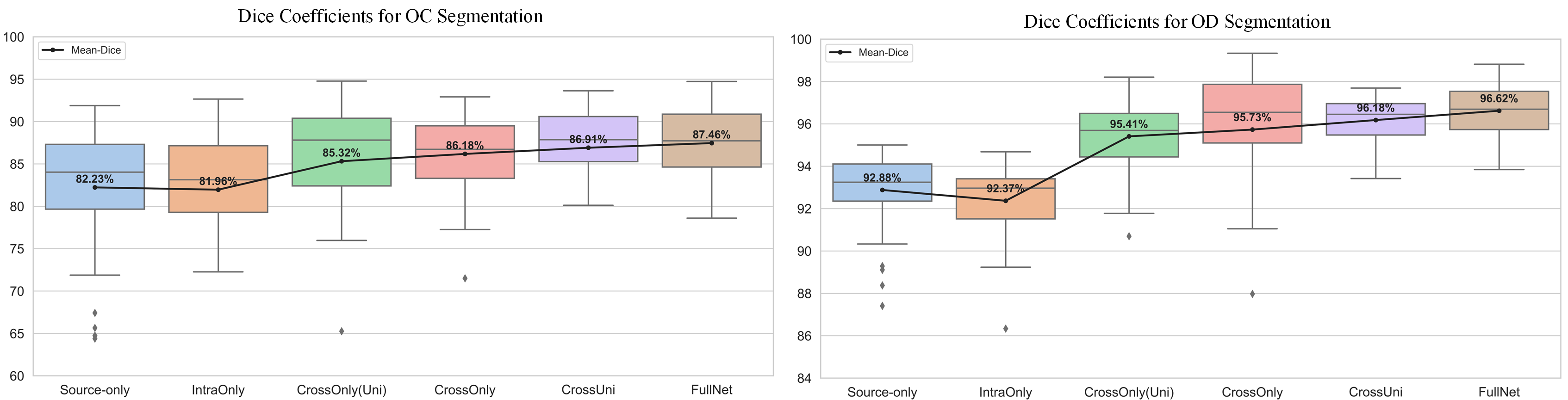}
  \caption{Boxplot of the OC/OD segmentation results produced by the different components in DDS\nobreakdash-UDA for the experiment on the Fundus dataset ``Domain 1".}
  \label{boxplot}
\end{figure*}

The quantitative analysis results on Dice and 95\% Hausdorff distance (HD) are shown in Table \ref{tab_Fundus}. We can clearly see that ``Target\nobreakdash-only" significantly outperformed ``Source\nobreakdash-only", which demonstrates the existence of domain shift between the source and target domains. 
The lower part of Table \ref{tab_Fundus} demonstrates that all the competitive methods achieved better performance than ``Source\nobreakdash-only". Taking the experiment in ``Domain 1" as an example, FDA, DoCR and CoFo achieved 84.27\%/94.89\%, 85.33\%/95.87\%, and 85.55\%/95.32\% on the Dice coefficient for the OC/OD segmentation, respectively, which were higher than those of ``Source\nobreakdash-only". TriLA achieved a suboptimal results of 86.69\%/96.21\% on this experiment, which were 0.77\%/0.41\% lower than our method. As for the Dice coefficient, our method obtained the best results in the OC segmentation on all the four experiments, and slightly lower results than TriLA in the OD segmentation on ``Domain 2". In particular,  compared with ``Source\nobreakdash-only", our method increased the performance of joint OC/OD segmentation by 5.23\%/3.74\%, 5.33\%/3.11\%, 5.86\%/3.67\% and 8.23\%/5.61\%, respectively, demonstrating the superiority in eliminating domain adaptation. As for computational complexity, we acknowledge a practical limitation that the proposed approach involves several hyperparameters that must be specified prior to training, which may increase the tuning effort and reduce out-of-the-box applicability across other target domains. Nevertheless, the method does not introduce additional learnable parameters beyond the segmentation backbone, and therefore the overall training time remains comparable under the same experimental setting. The visualization results of all the methods were presented in Fig. \ref{Fundus_visual}. It is clear that our method produced segmentation results that were more consistent with the ground truth, especially in the ``Domain 2" experiment, and the other methods produced either over\nobreakdash- or under\nobreakdash-segmentation in various parts. This again indicates the effectiveness of our method in mitigating domain shift.

\begin{figure}[h]
  \centering
  \includegraphics[width=0.5\textwidth]{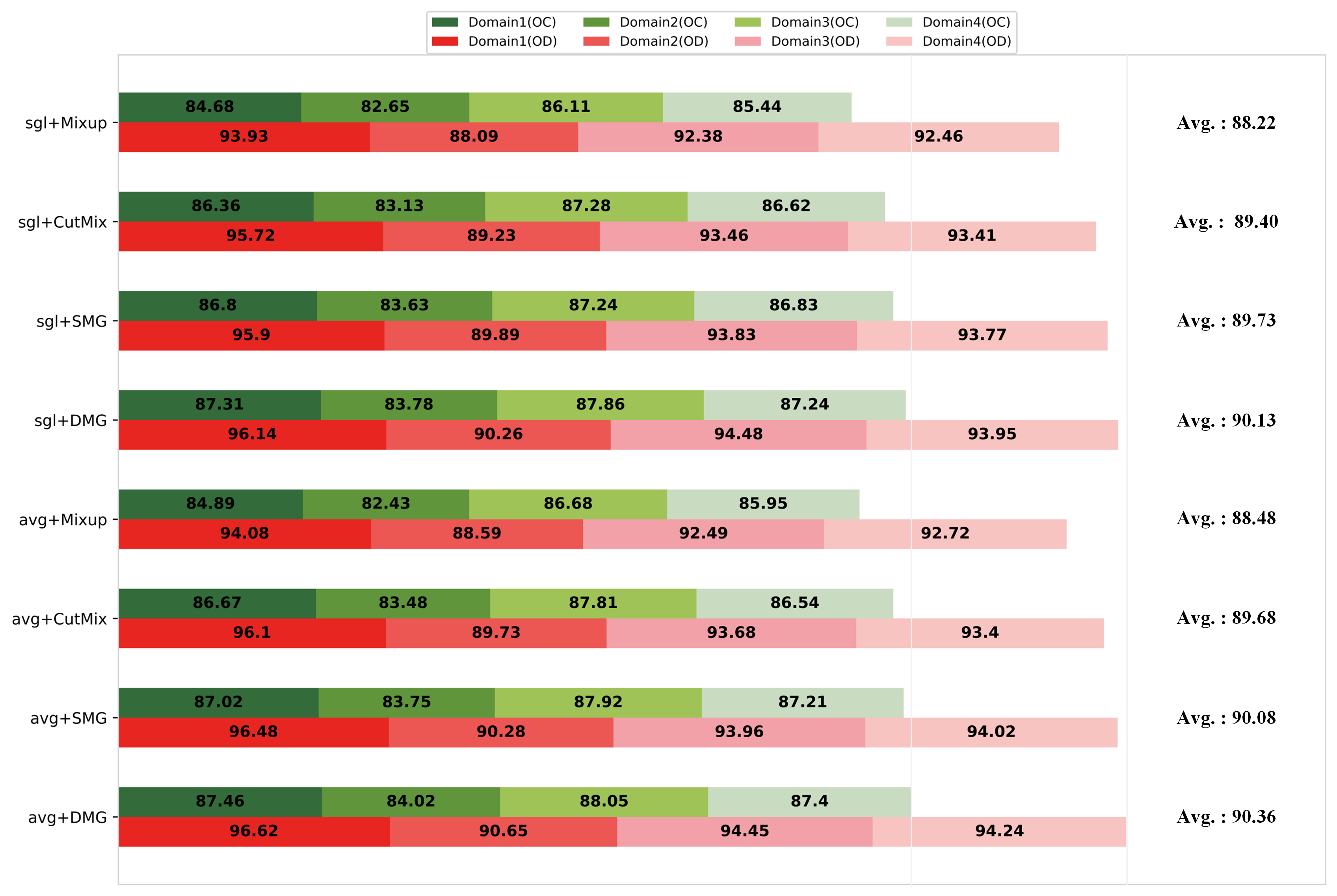}
  \caption{Accumulative bar diagram for OC/OD segmentation performance of different intra\nobreakdash- and cross\nobreakdash-domain data processing techniques on the Fundus dataset.}
  \label{conti_bar}
\end{figure}

\begin{figure*}[h]
  \centering
  \includegraphics[width=\textwidth]{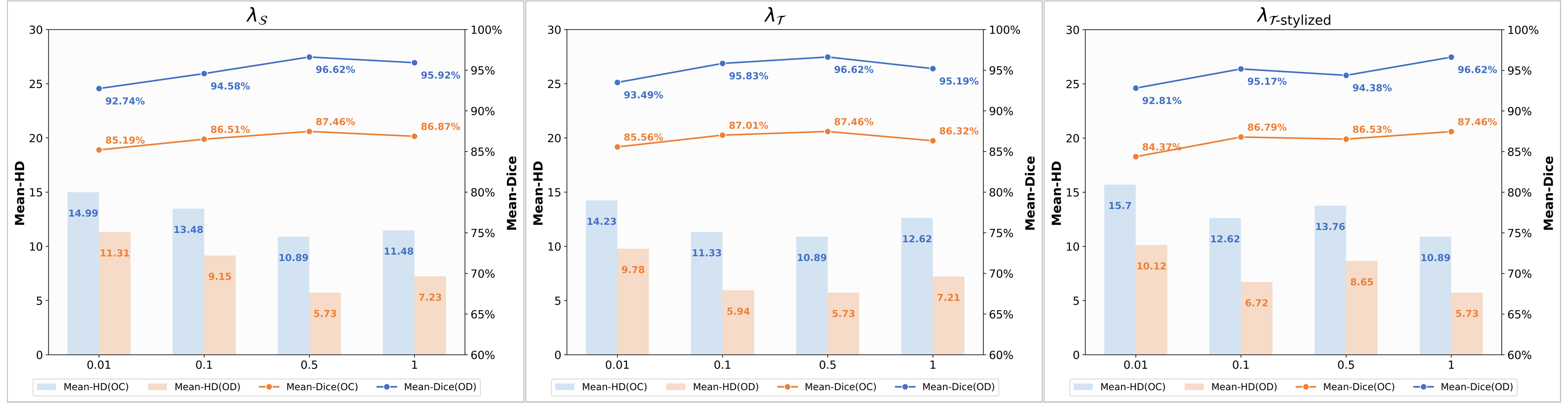}
  \caption{The influence of the tunable parameters, i.e., $\lambda_\mathcal{S}$, $\lambda_\mathcal{T}$, and $\lambda_{\mathcal{T}\text{-stylized}}$, on the ``Domain 1" scenario of the Fundus dataset.}
  \label{bar_dash}
\end{figure*}

\begin{table*}[ht]
\renewcommand{\arraystretch}{1.2}
\belowrulesep=0pt
\aboverulesep=0pt
\centering
    \caption{Dice results on the RIGA+ dataset produced by different components in DDS\nobreakdash-UDA. ``Avg." indicates the average of the results produced by the corresponding method under six experiments.}
    \label{tab:performance_comparison}
    \resizebox{\textwidth}{!}{%
    \begin{tabular}{lcc|cc|cc|cc|cc|cc|c}
    \toprule
    \multicolumn{1}{c|}{Source Domain} & \multicolumn{6}{c|}{Domain1 (BinRushed)} & \multicolumn{6}{c|}{Domain2 (Magrabia)} & \multirow{3}{*}{Avg.} \\
    \cline{1-13}
    \multicolumn{1}{c|}{Target Domain (MESSIDOR)}& \multicolumn{2}{c|}{Base1} & \multicolumn{2}{c|}{Base2} & \multicolumn{2}{c|}{Base3} & \multicolumn{2}{c|}{Base1} & \multicolumn{2}{c|}{Base2} & \multicolumn{2}{c|}{Base3} & \\
    \cline{1-13}
    \multicolumn{1}{c|}{Method} & OC & OD & OC & OD & OC & OD & OC & OD & OC & OD & OC & OD \\
    \midrule
    \multicolumn{1}{c|}{Source-only} & 83.01 & 94.08 & 83.76 & 94.96 & 83.81 & 93.69 & 83.13 & 93.87 & 84.77 & 94.01 & 83.28 & 94.06 & 88.87 \\
    \multicolumn{1}{c|}{IntraOnly} & 82.66 & 92.39 & 84.58 & 95.28 & 83.31 & 92.48 & 82.98 & 93.74 & 84.36 & 93.68 & 83.13 & 94.01 & 88.55 \\
    \multicolumn{1}{c|}{CrossOnly (Uni)} & 85.61 & 94.53 & 85.68 & 95.67 & 85.07 & 94.77 & 84.32 & 95.28 & 86.30 & 95.36 & 86.27 & 94.89 & 90.31 \\
    \multicolumn{1}{c|}{CrossOnly} & 87.25 & 95.21 & 86.12 & 95.82 & 85.82 & 94.91 & 84.58 & 95.87 & 86.98 & 95.75 & 86.93 & 95.16 & 90.87 \\
    \multicolumn{1}{c|}{CrossUni} & 86.57 & 95.48 & \textbf{86.91} & 96.19 & 86.21 & 95.17 & 84.85 & 95.63 & 86.92 & 96.03 & 86.98 & 95.26 & 91.02 \\
    \multicolumn{1}{c|}{FullNet (Ours)} & \textbf{87.40} & \textbf{95.90} & 86.44 & \textbf{96.20} & \textbf{86.64} & \textbf{95.31} & \textbf{85.28} & \textbf{96.00} & \textbf{87.30} & \textbf{96.34} & \textbf{87.14} & \textbf{95.71} & \textbf{91.31} \\
    \bottomrule
    \end{tabular}}
\end{table*}

\begin{figure}[h]
  \centering
  \includegraphics[width=0.4\textwidth]{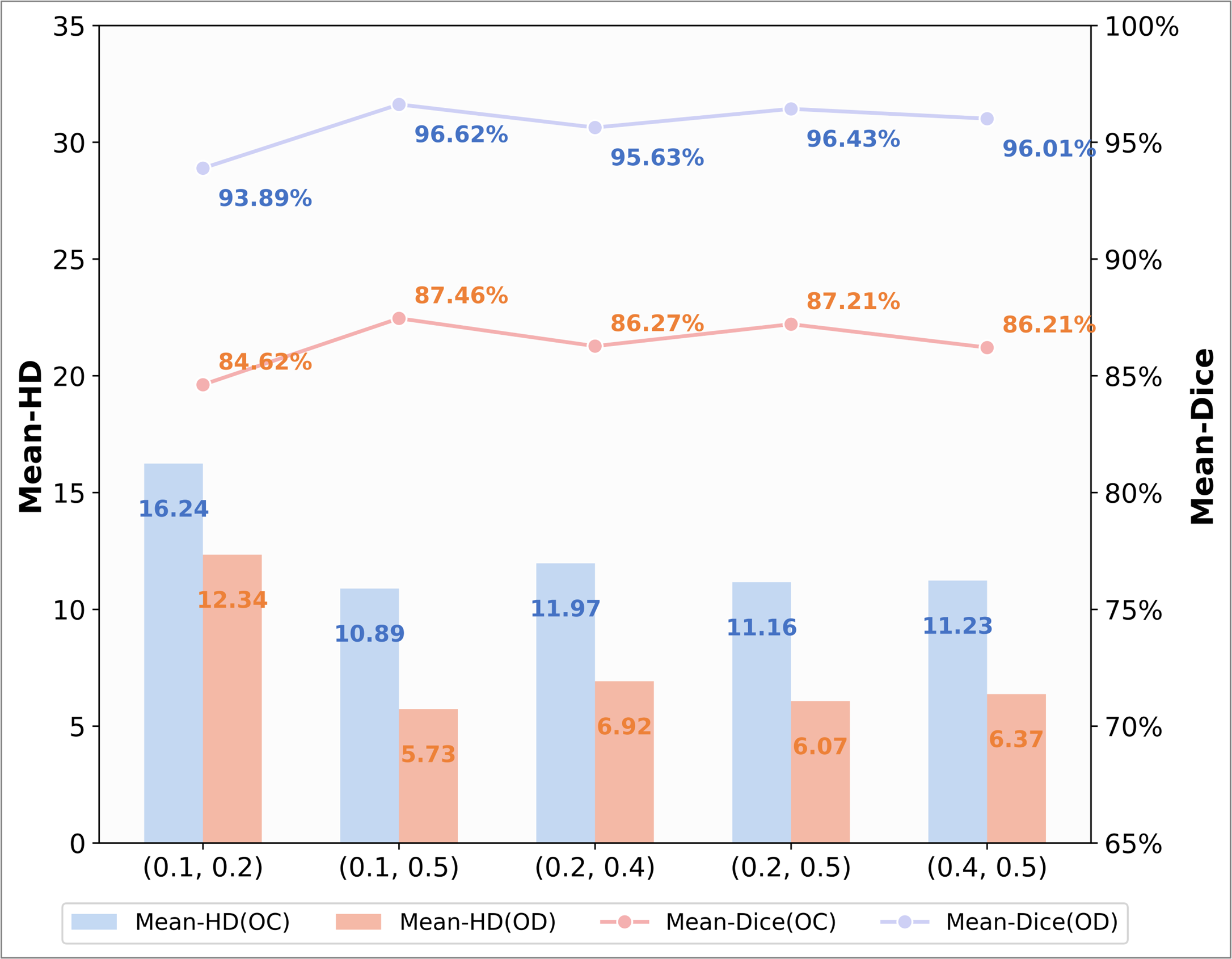}
  \caption{Impact of the interval bounds of the parameter $\lambda_k$ on the model performance.}
  \label{threshold}
\end{figure}

\subsection{Ablation Analysis}
In order to evaluate the effectiveness of the three key components in DDS\nobreakdash-UDA, i.e., the intra\nobreakdash-domain pseudo label learning component, the cross\nobreakdash-domain consistency component, and the bi\nobreakdash-directional copy\nobreakdash-paste component, we designed an ablation study with the following combinations of these components.
\begin{itemize}
    \item IntraOnly:  Retaining only the intra\nobreakdash-domain pseudo label learning, where the spectral\nobreakdash-stylized target data and raw target data were processed without cross\nobreakdash-domain interaction. Source domain data were utilized solely during pre\nobreakdash-training.
    \item CrossOnly: Replacing the raw target data with the spectral\nobreakdash-stylized target data in the teacher\nobreakdash-student training loop and adjusting the prediction loss calculation to account for style\nobreakdash-transferred inputs correspondingly.
    \item CrossUni: Performing unidirectional cross\nobreakdash-domain fusion using fixed, non\nobreakdash-inverse masks for each source\nobreakdash-target image pair.
    \item CrossOnly (Uni): Combining the ``CrossOnly" setting with unidirectional cross\nobreakdash-domain fusion.
    \item FullNet (ours): Combining all the three components together.
\end{itemize}
% and Table. \ref{tab:performance_comparison}
\par Without loss of generality, we performed this ablation study on the ``Domain 1" experiment. The results of the OC/OD segmentation are shown in Fig. \ref{boxplot}. We included the ``Source\nobreakdash-only" to well measure the contribution of each component of our DDS\nobreakdash-UDA. From Fig. \ref{boxplot}, it can be clearly seen that ``Source\nobreakdash-only" produced a certain number of anomalies below the segmented OC and OD regions. It suggested that there is severe domain shift between the source and target domains, leading to the poor robustness of the model trained on the source domain data only. Compared to ``Source\nobreakdash-only", ``IntraOnly" exhibited worse segmentation performance due to the lack of cross\nobreakdash-domain information exchange. And the last four methods that incorporate cross\nobreakdash-domain information exchange surpassed ``Source\nobreakdash-only" in the OC segmentation by 3.09\%, 3.95\%, 4.68\%, 5.23\%, respectively. Statistical analysis using paired t-tests confirmed that all these differences are statistically significant (p-values: $1.06 \times 10^{-2}$, $4.84 \times 10^{-4}$, $1.13 \times 10^{-5}$, $1.04 \times 10^{-6}$, respectively).

\begin{table}[h]
\renewcommand{\arraystretch}{1.5}
\centering
\caption{Statistics from the RIGA+ dataset used in this study.}
\resizebox{\linewidth}{!}{
\begin{tabular}{l|cc|ccc}
\hline
\multirow{2}{*}{\textbf{Domain}} & \multicolumn{2}{c|}{\textbf{Source}} & \multicolumn{3}{c}{\textbf{Target (MESSIDOR)}} \\
\cline{2-6}
 & BinRushed & Magrabia & BASE1 & BASE2 & BASE3 \\
\hline
\textbf{Training} & 195 & 95 & 138 & 118 & 106 \\
\textbf{Testing} & 0 & 0 & 35 & 30 & 27 \\
\textbf{Unlabeled} & -- & -- & 227 & 238 & 252 \\
\hline
\textbf{Overall} & 195 & 95 & 400 & 386 & 385 \\
\hline
\end{tabular}
}
\label{stas_riga}
\end{table}

\par Compared to ``CrossUni" and ``FullNet", which incorporate the intra\nobreakdash-domain pseudo label learning, ``CrossOnly (Uni)" and ``CrossOnly" exhibit a higher frequency of anomalies, reflecting the fact that predictions generated by relying solely on the bi\nobreakdash-directional cross\nobreakdash-domain consistency are not sufficiently stable. On the contrary, the use of the intra\nobreakdash-domain pseudo label learning effectively mitigates the inter\nobreakdash-domain bias and enhances the model's adaptability to anomalies and variations in different domain characteristics. Specifically, as can be seen in Fig. \ref{boxplot}, ``FullNet" and ``CrossUni" outperforms ``CrossOnly" and ``CrossOnly (Uni)" on the mean\nobreakdash-dice metric for the OD segmentation by 0.89\% and 0.77\%, respectively. Statistical significance is confirmed through paired t-tests (p-values of 2.19e-4 and 7.99e-3). In order to analyze the effectiveness of the bi\nobreakdash-directional copy\nobreakdash-paste, we find that the mean\nobreakdash-dice of the former is higher than that of the latter in the OC and OD segmentations by comparing ``FullNet" and ``CrossUni", respectively. The 95\% confidence intervals for these improvements are [0.06, 1.33] and [0.13, 1.06], demonstrating that the observed improvements are both statistically significant. The improvement in segmentation performance implies that the bi\nobreakdash-directional copy\nobreakdash-paste captures the fine\nobreakdash-grained information of images better than uni\nobreakdash-directional fusion.

\begingroup
\setlength{\textfloatsep}{12pt}
\setlength{\abovecaptionskip}{8pt}
\setlength{\belowcaptionskip}{8pt}
\renewcommand{\arraystretch}{1.3}

\begin{table}[h]
\centering
\caption{Performance variation with different values of $d_{min}$ and $d_{max}$.}
\label{addition sensitivity}

\begin{tabular}{c!{\vrule width 1pt}ccc}
\toprule[1.6pt]
\diagcell{$d_{\min}$}{$d_{\max}$} & 2 & 3 & 5 \\
\midrule[0.8pt]
0.1 & 93.91/84.52 & 94.38/85.21 & 96.03/86.11 \\
1   & 95.32/85.98 & 96.18/86.50 & \textbf{96.62/87.46} \\
3   & \textemdash & 96.48/87.02 & 96.47/87.18 \\
\bottomrule[1.6pt]
\end{tabular}
\end{table}
\endgroup

\renewcommand{\mstd}[2]{\ensuremath{#1_{\scriptscriptstyle #2}}}

\begin{table*}[htb]
\centering
\scriptsize
\renewcommand{\arraystretch}{1.22}
\setlength{\tabcolsep}{2.2pt}
\belowrulesep=0pt
\aboverulesep=0pt
\caption{Comparisons with other methods on the RIGA+ dataset. Results are reported as mean$_{\mathrm{std}}$ over three runs.}
\label{tab:riga_plus_all}

% ===================== Domain 1 =====================
\resizebox{\textwidth}{!}{
\begin{tabular}{c|cccc|cccc|cccc}
\toprule
\multirow{2}{*}{\parbox{2.2cm}{\centering Method}} &
\multicolumn{12}{c}{Domain1 (BinRushed)} \\
\cmidrule(lr){2-13}
& \multicolumn{4}{c|}{Base1} & \multicolumn{4}{c|}{Base2} & \multicolumn{4}{c}{Base3} \\
\cmidrule(lr){2-5}\cmidrule(lr){6-9}\cmidrule(lr){10-13}
& $HD_{OD}\downarrow$ & $Dice_{OD}\uparrow$ & $HD_{OC}\downarrow$ & $Dice_{OC}\uparrow$
& $HD_{OD}\downarrow$ & $Dice_{OD}\uparrow$ & $HD_{OC}\downarrow$ & $Dice_{OC}\uparrow$
& $HD_{OD}\downarrow$ & $Dice_{OD}\uparrow$ & $HD_{OC}\downarrow$ & $Dice_{OC}\uparrow$ \\
\midrule

Source-only
& \mstd{8.21}{0.42} & \mstd{94.08}{0.26} & \mstd{10.49}{0.21} & \mstd{83.01}{0.38}
& \mstd{8.13}{0.50} & \mstd{94.96}{0.54} & \mstd{9.94}{0.38} & \mstd{83.76}{0.36}
& \mstd{8.77}{0.46} & \mstd{93.69}{0.60} & \mstd{10.64}{0.13} & \mstd{83.81}{0.39} \\

Target-only
& \mstd{4.98}{0.21} & \mstd{95.78}{0.23} & \mstd{6.03}{0.34} & \mstd{86.91}{0.25}
& \mstd{4.87}{0.10} & \mstd{96.31}{0.11} & \mstd{7.36}{0.21} & \mstd{86.88}{0.14}
& \mstd{5.35}{0.23} & \mstd{95.03}{0.24} & \mstd{6.53}{0.26} & \mstd{86.17}{0.16} \\

\midrule

FDA \cite{DBLP:conf/cvpr/0001S20}
& \mstd{7.82}{0.29} & \mstd{94.98}{0.42} & \mstd{9.73}{0.27} & \mstd{83.36}{0.47}
& \mstd{7.47}{0.47} & \mstd{95.37}{0.20} & \mstd{10.01}{0.39} & \mstd{83.77}{0.46}
& \mstd{8.38}{0.43} & \mstd{94.60}{0.54} & \mstd{10.45}{0.62} & \mstd{84.39}{0.45} \\

DoCR \cite{DBLP:conf/miccai/HuLX22}
& \mstd{6.17}{0.41} & \mstd{95.34}{0.29} & \mstd{8.13}{0.48} & \mstd{85.73}{0.41}
& \mstd{6.39}{0.32} & \mstd{95.75}{0.26} & \mstd{8.74}{0.51} & \mstd{83.80}{0.47}
& \mstd{6.54}{0.43} & \mstd{94.82}{0.31} & \mstd{7.84}{0.46} & \mstd{86.62}{0.38} \\

DAFormer \cite{hoyer2022daformer}
& \mstd{6.05}{0.40} & \mstd{95.45}{0.48} & \mstd{8.01}{0.37} & \mstd{85.95}{0.39}
& \mstd{6.20}{0.31} & \mstd{95.80}{0.65} & \mstd{8.60}{0.20} & \mstd{84.40}{0.94}
& \mstd{6.35}{0.42} & \mstd{94.95}{0.79} & \mstd{7.70}{0.65} & \mstd{86.10}{0.37} \\

CoFo \cite{DBLP:conf/isbi/HuyHNDBT22}
& \mstd{7.50}{0.28} & \mstd{95.24}{0.70} & \mstd{8.64}{0.52} & \mstd{84.20}{0.84}
& \mstd{6.92}{0.35} & \mstd{95.04}{0.41} & \mstd{9.57}{0.57} & \mstd{85.10}{0.52}
& \mstd{7.90}{0.40} & \mstd{94.63}{0.63} & \mstd{9.86}{0.58} & \mstd{85.00}{0.42} \\

BEAL \cite{DBLP:journals/cmpb/LiuPSS22}
& \mstd{7.68}{0.29} & \mstd{95.33}{0.69} & \mstd{9.21}{0.55} & \mstd{85.08}{0.93}
& \mstd{6.50}{0.22} & \mstd{94.91}{0.43} & \mstd{9.22}{0.46} & \mstd{84.99}{0.74}
& \mstd{7.39}{0.47} & \mstd{94.23}{0.35} & \mstd{9.18}{0.59} & \mstd{84.38}{0.55} \\

ODADA \cite{DBLP:journals/mia/SunDX22}
& \mstd{6.55}{0.23} & \mstd{95.32}{0.29} & \mstd{7.84}{0.12} & \mstd{85.11}{0.44}
& \mstd{5.48}{0.17} & \mstd{95.39}{0.28} & \mstd{7.84}{0.16} & \mstd{85.61}{0.42}
& \mstd{6.83}{0.24} & \mstd{94.61}{0.32} & \mstd{8.49}{0.20} & \mstd{84.62}{0.46} \\

TIST \cite{DBLP:conf/miccai/GhamsarianTMWZSS23}
& \mstd{6.54}{0.43} & \mstd{95.21}{0.40} & \mstd{7.98}{0.47} & \mstd{83.31}{0.48}
& \mstd{7.22}{0.36} & \mstd{95.18}{0.50} & \mstd{8.97}{0.53} & \mstd{84.29}{0.46}
& \mstd{7.97}{0.10} & \mstd{94.66}{0.52} & \mstd{9.36}{0.55} & \mstd{85.97}{0.41} \\

MA\nobreakdash-UDA \cite{ji2023unsupervised}
& \mstd{5.85}{0.28} & \mstd{95.55}{0.37} & \mstd{7.30}{0.13} & \mstd{86.40}{0.56}
& \mstd{5.60}{0.16} & \mstd{95.88}{0.14} & \mstd{7.45}{0.24} & \mstd{86.10}{0.67}
& \mstd{6.00}{0.49} & \mstd{95.05}{0.26} & \mstd{7.05}{0.41} & \mstd{86.35}{0.35} \\

TriLA \cite{DBLP:journals/titb/ChenPYWX24}
& \mstd{5.18}{0.25} & \mstd{95.41}{0.36} & \mstd{7.68}{0.14} & \mstd{86.15}{0.56}
& \mstd{5.18}{0.45} & \mstd{95.70}{0.35} & \mstd{7.59}{0.23} & \mstd{87.01}{0.65}
& \mstd{6.26}{0.19} & \mstd{94.54}{0.28} & \mstd{7.15}{0.11} & \mstd{86.25}{0.26} \\

\midrule
\textbf{Ours}
& \textbf{\mstd{4.44}{0.18}} & \textbf{\mstd{95.90}{0.34}} & \textbf{\mstd{6.12}{0.16}} & \textbf{\mstd{87.40}{0.43}}
& \textbf{\mstd{5.10}{0.06}} & \textbf{\mstd{96.20}{0.22}} & \textbf{\mstd{7.71}{0.23}} & \textbf{\mstd{86.84}{0.24}}
& \textbf{\mstd{5.50}{0.22}} & \textbf{\mstd{95.31}{0.25}} & \textbf{\mstd{6.75}{0.20}} & \textbf{\mstd{86.64}{0.25}} \\
\bottomrule
\end{tabular}
}

\vspace{2mm}

% ===================== Domain 2 =====================
\resizebox{\textwidth}{!}{
\begin{tabular}{c|cccc|cccc|cccc}
\toprule
\multirow{2}{*}{\parbox{2.2cm}{\centering Method}} &
\multicolumn{12}{c}{Domain2 (Magrabia)} \\
\cmidrule(lr){2-13}
& \multicolumn{4}{c|}{Base1} & \multicolumn{4}{c|}{Base2} & \multicolumn{4}{c}{Base3} \\
\cmidrule(lr){2-5}\cmidrule(lr){6-9}\cmidrule(lr){10-13}
& $HD_{OD}\downarrow$ & $Dice_{OD}\uparrow$ & $HD_{OC}\downarrow$ & $Dice_{OC}\uparrow$
& $HD_{OD}\downarrow$ & $Dice_{OD}\uparrow$ & $HD_{OC}\downarrow$ & $Dice_{OC}\uparrow$
& $HD_{OD}\downarrow$ & $Dice_{OD}\uparrow$ & $HD_{OC}\downarrow$ & $Dice_{OC}\uparrow$ \\
\midrule

Source-only
& \mstd{8.34}{0.23} & \mstd{93.87}{0.48} & \mstd{10.57}{0.12} & \mstd{83.13}{0.50}
& \mstd{8.81}{0.17} & \mstd{94.01}{0.77} & \mstd{9.97}{0.29} & \mstd{84.77}{0.44}
& \mstd{8.42}{0.34} & \mstd{94.06}{0.66} & \mstd{11.68}{0.30} & \mstd{83.28}{0.89} \\

Target-only
& \mstd{5.89}{0.25} & \mstd{96.18}{0.62} & \mstd{6.61}{0.17} & \mstd{85.67}{0.38}
& \mstd{4.98}{0.31} & \mstd{96.74}{0.40} & \mstd{5.33}{0.22} & \mstd{87.59}{0.33}
& \mstd{5.22}{0.32} & \mstd{95.41}{0.33} & \mstd{7.89}{0.14} & \mstd{87.26}{0.34} \\

\midrule

FDA \cite{DBLP:conf/cvpr/0001S20}
& \mstd{7.92}{0.50} & \mstd{94.91}{0.23} & \mstd{9.49}{0.26} & \mstd{83.84}{0.66}
& \mstd{7.86}{0.49} & \mstd{95.18}{0.11} & \mstd{9.26}{0.15} & \mstd{85.08}{0.22}
& \mstd{7.27}{0.31} & \mstd{94.69}{0.43} & \mstd{10.28}{0.26} & \mstd{83.98}{0.35} \\

DoCR \cite{DBLP:conf/miccai/HuLX22}
& \mstd{6.13}{0.38} & \mstd{94.98}{0.33} & \mstd{7.84}{0.29} & \mstd{84.34}{0.45}
& \mstd{5.70}{0.36} & \mstd{95.32}{0.29} & \mstd{7.47}{0.17} & \mstd{85.18}{0.53}
& \mstd{6.99}{0.42} & \mstd{94.73}{0.32} & \mstd{8.21}{0.22} & \mstd{86.61}{0.69} \\

DAFormer \cite{hoyer2022daformer}
& \mstd{6.05}{0.37} & \mstd{95.21}{0.50} & \mstd{7.17}{0.47} & \mstd{84.61}{0.54}
& \mstd{5.92}{0.35} & \mstd{95.55}{0.48} & \mstd{7.34}{0.48} & \mstd{86.15}{0.40}
& \mstd{6.52}{0.41} & \mstd{95.13}{0.39} & \mstd{7.93}{0.51} & \mstd{85.93}{0.31} \\

CoFo \cite{DBLP:conf/isbi/HuyHNDBT22}
& \mstd{6.18}{0.37} & \mstd{96.01}{0.28} & \mstd{7.31}{0.46} & \mstd{85.22}{0.53}
& \mstd{6.79}{0.40} & \mstd{94.48}{0.34} & \mstd{8.21}{0.52} & \mstd{84.83}{0.64}
& \mstd{6.19}{0.37} & \mstd{94.94}{0.32} & \mstd{9.27}{0.58} & \mstd{84.84}{0.24} \\

BEAL \cite{DBLP:journals/cmpb/LiuPSS22}
& \mstd{7.89}{0.29} & \mstd{94.50}{0.54} & \mstd{8.57}{0.34} & \mstd{83.41}{0.57}
& \mstd{6.55}{0.41} & \mstd{95.19}{0.41} & \mstd{8.12}{0.22} & \mstd{86.95}{0.68}
& \mstd{5.13}{0.32} & \mstd{94.72}{0.53} & \mstd{8.94}{0.46} & \mstd{85.20}{0.53} \\

ODADA \cite{DBLP:journals/mia/SunDX22}
& \mstd{6.84}{0.22} & \mstd{94.93}{0.31} & \mstd{9.56}{0.31} & \mstd{85.55}{0.53}
& \mstd{6.69}{0.41} & \mstd{95.22}{1.19} & \mstd{7.95}{0.30} & \mstd{86.28}{0.71}
& \mstd{6.10}{0.39} & \mstd{94.85}{0.82} & \mstd{8.90}{0.27} & \mstd{85.80}{1.02} \\

TIST \cite{DBLP:conf/miccai/GhamsarianTMWZSS23}
& \mstd{6.89}{0.23} & \mstd{93.67}{0.35} & \mstd{8.19}{0.32} & \mstd{84.18}{0.66}
& \mstd{6.61}{0.11} & \mstd{95.43}{0.28} & \mstd{7.12}{0.15} & \mstd{85.37}{0.83}
& \mstd{6.76}{0.29} & \mstd{94.73}{0.32} & \mstd{9.44}{0.29} & \mstd{86.37}{0.35} \\

MA\nobreakdash-UDA \cite{ji2023unsupervised}
& \mstd{5.91}{0.34} & \mstd{95.17}{0.17} & \mstd{7.54}{0.13} & \mstd{85.02}{0.41}
& \mstd{6.12}{0.33} & \mstd{95.35}{0.25} & \mstd{6.49}{0.31} & \mstd{88.24}{0.62}
& \mstd{6.45}{0.26} & \mstd{94.68}{0.27} & \mstd{8.86}{0.28} & \mstd{85.78}{0.39} \\

TriLA \cite{DBLP:journals/titb/ChenPYWX24} 
& \mstd{5.73}{0.16} & \mstd{95.62}{0.25} & \mstd{6.58}{0.21} & \mstd{84.75}{0.74}
& \mstd{5.26}{0.23} & \mstd{95.77}{0.24} & \mstd{6.13}{0.29} & \mstd{88.90}{0.61}
& \mstd{5.79}{0.27} & \mstd{95.00}{0.27} & \mstd{8.51}{0.33} & \mstd{85.66}{0.49} \\

\midrule
\textbf{Ours}
& \textbf{\mstd{5.57}{0.15}} & \textbf{\mstd{96.00}{0.23}} & \textbf{\mstd{6.20}{0.08}} & \textbf{\mstd{85.28}{0.27}}
& \textbf{\mstd{5.04}{0.23}} & \textbf{\mstd{96.34}{0.52}} & \textbf{\mstd{5.96}{0.16}} & \textbf{\mstd{87.30}{0.33}}
& \textbf{\mstd{5.42}{0.14}} & \textbf{\mstd{95.71}{0.24}} & \textbf{\mstd{7.37}{0.24}} & \textbf{\mstd{87.14}{0.43}} \\
\bottomrule
\end{tabular}
}  

\end{table*}

\subsection{Impact of Intra- and Cross-Domain Data Processing Techniques}
To investigate the effect of the intra\nobreakdash- and cross\nobreakdash-domain data processing on the results, we performed additional experiments by using different data processing techniques. The experimental results for multiple variants of the proposed method are shown in Fig. \ref{conti_bar}. Specifically, the intra\nobreakdash-domain data processing refers to how the stylized target images are generated.  We analyzed the batch\nobreakdash-average spectrum (abbr. avg) and single\nobreakdash-image spectrum (sgl) approaches in Fig. \ref{conti_bar}. As for mask generation in the cross\nobreakdash-domain data processing, we compared four categories of methods: Mixup \cite{mixup}, CutMix \cite{cutmix}, static mask generator (SMG) and dynamic mask generator (DMG), where SMG refers to the case where the corresponding $d$\nobreakdash-value in (\ref{ccc}) is static. Notably, when $d$ is static, the generated masks preserve randomness. From Fig. \ref{conti_bar}, it can be clearly seen that in both the ``sgl" and ``avg" cases, segmentation results of ``DMG" in the four experimental results are higher than those of ``SMG". This advantage is even more pronounced compared to ``CutMix", as fixing the mask shape often leads to an over\nobreakdash-reliance on a particular mixing pattern in the model, which produces poor generalization results when tested in the target domain. This demonstrates that the introduction of DMG can effectively enhance the model's accuracy. In the comparison between ``avg" and ``sgl", in most cases the mean\nobreakdash-dice metric of ``avg" is slightly higher than that of ``sgl", and even in the OD segmentation results of ``Domain 3", the value of the mean\nobreakdash-dice evaluation metric of ``avg" is higher than that of ``sgl" by 0.57\%. This indicates that the batch average spectrum approach can better improve the robustness of the model while reducing the amplitude instability brought by a single image.

\begin{figure*}[h]
  \centering
  \includegraphics[width=\textwidth,height=0.8\textheight,keepaspectratio]{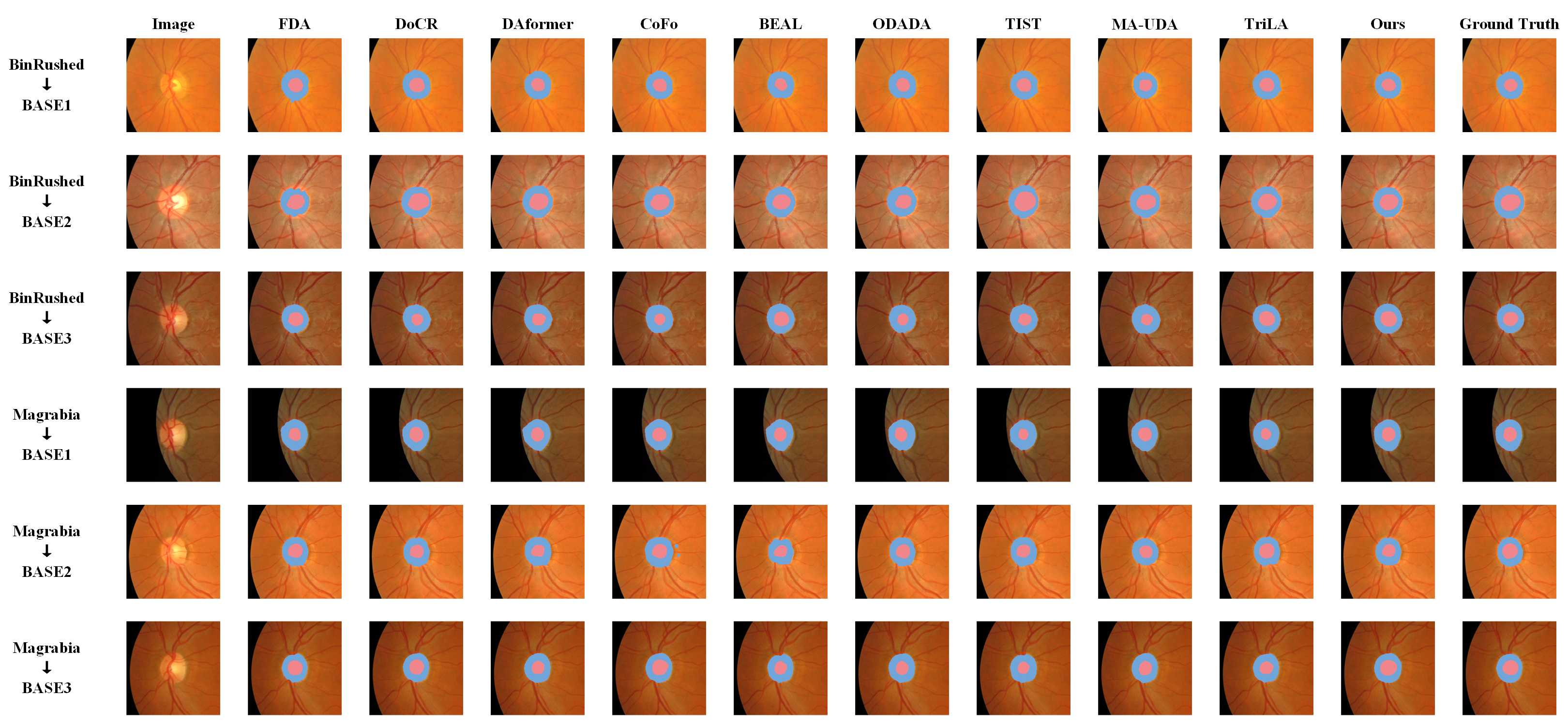}
  \caption{The visualization results produced by different methods on the RIGA+ dataset. Red and blue colors denote OC and OD, respectively.}
  \label{RIGA+_visual}
\end{figure*}

\subsection{Influence of Hyperparameters}
In our experiments, In our experiments, we conducted a systematic sensitivity analysis to assess the impact of key hyperparameters and determine their appropriate settings. For the loss function weight parameters $\lambda_\mathcal{S}$, $\lambda_\mathcal{T}$ and $\lambda_{\mathcal{T}\text{-stylized}}$, we varied each weight over {0.01, 0.1, 0.5, 1.0} in a one\nobreakdash-factor\nobreakdash-at\nobreakdash-a\nobreakdash-time study, holding all other variables fixed. The results on the ``Domain 1" experiment of the Fundus dataset are shown in Fig. \ref{bar_dash}. It showed that $\lambda_{\mathcal{T}\text{-stylized}}$ exhibits the most significant variation in the Dice metric, with the highest and lowest values differing by 3.09$\%$ and 3.81$\%$ for the OC and OD segmentations, respectively. This indicated that the proportion of the loss function in the domain\nobreakdash-specific pseudo label learning determines the quality of pseudo label supervision, thereby affecting the robustness of the model. In all settings, $\lambda_\mathcal{S}$ and $\lambda_\mathcal{T}$ reached optimal performance at 0.5, and deviation from this optimum led to a decline in the evaluation metrics in both increasing and decreasing directions. This balance suggests that lower weights do not adequately perform anatomical constraints, while higher weights inhibit the expression of information in the other domain.
\par In addition, the parameter $\lambda_k$ in Eq. \ref{ccc} was randomly generated within a specified interval during the training process. To evaluate the impact of the interval bounds on the results, we conducted further experiments using the Fundus dataset in the “Domain 1” experiment, with the results illustrated in Fig. \ref{threshold}. Since the generated masks were bidirectional in subsequent stages, experiments were not conducted for intervals above 0.5. The analysis reveals that the interval (0.1, 0.2) resulted in poorer performance, likely due to the smaller values generated within this range, which may have restricted the information exchange between the source and target domains. The results show an increasing trend as the interval expands, with the optimal performance observed at the (0.1, 0.5) interval, where both OC and OD achieve the highest Dice coefficients.
\par We further evaluated the sensitivity to the attenuation factor $d$ in Eq. \ref{di}, which modulates the spectral decay in the dynamic mask generator and therefore controls the spatial granularity of the resulting masks. We varied the schedule endpoints $(d_{min},d_{max})$ in Eq. \ref{di} and reported the corresponding results on the Fundus “Domain 1” experiment in Table. \ref{addition sensitivity}. The performance was relatively stable across a broad range of settings, and the best results were obtained when $d_{min}$ and $d_{max}$ are set to 1 and 5. These observations suggested that the proposed method is not unduly sensitive to the precise choice of $d$, and a moderate coarse\nobreakdash-to\nobreakdash-fine schedule suffices to achieve robust performance.

\subsection{Evaluation of Model Generalizability}
To assess the generalizability of our model, we conducted comprehensive experiments on another fundus dataset, RIGA+ \cite{RIGA,RIGAPlus}, for further validation. Table \ref{stas_riga} shows the relevant information for this dataset.
\subsubsection{RIGA+} The multi\nobreakdash-domain dataset RIGA+ consisted of the BinRushed dataset, the Magrabia dataset, and the MESSIDOR dataset which contained data from three medical centers, namely BASE1, BASE2, and BASE3. Each labeled image in the RIGA+ dataset provided multiple annotations, and we only use the annotations of rater No. 1 to train and evaluate the network. All images were center\nobreakdash-cropped and resized to 256$\times$256. We utilized the BinRushed and Magrabia datasets as two source domains, while BASE1, BASE2, and BASE3 were designated as three target domains. In total, this configuration produced 6 UDA scenarios, with each scenario involving a unique source\nobreakdash-target pair.

\subsubsection{Experiments on the RIGA+ dataset}
To further validate the proposed DDS\nobreakdash-UDA framework, we presented comparison results on the RIGA+ dataset. As listed in Table \ref{tab:riga_plus_all}, the domain shift on this dataset remained significant, with ``Source\nobreakdash-only" and ``Target\nobreakdash-only" having an average dice discrepancy of 1.81\%/2.91\% across the 6 domain adaptation experiments. Compared to the ``Source\nobreakdash-only" method, the existing methods showed improvements in both Dice and HD metrics with varying degrees. In terms of the Dice coefficient, the ``TriLA" method achieved excellent results in several experimental scenarios because of its proposed triple\nobreakdash-level alignment model. Compared to the ``TriLA" method, our proposed DDS\nobreakdash-UDA method exhibited only minor performance deficits in OD segmentation for the ``BinRushed$\to$Base2" and ``Magrabia$\to$Base2" experimental scenarios, with Dice scores lower by 0.37\% and 0.6\%, respectively. However, our method demonstrated superior performance in all other scenarios. Moreover, this advantage becomes even more pronounced when compared to the other competing methods. Compared with the transformer-based MA\nobreakdash-UDA, our method demonstrates a consistent improvement in terms of the Dice metric. In terms of HD metrics on OD segmentation, our method achieved 4.44, 5.10, 5.50 in the three experiments of the ``BinRushed$\to$MESSIDOR" experimental scenario, which were 0.49\%, 0.45\% and 0.49\% higher than those of the best method available. 
\par Notably, our method had a slight performance improvement on ``BinRushed$\to$Base1" and ``BinRushed$\to$Base3" compared to the ``Target\nobreakdash-only" method that used the target domain labeling. This showed that our method can effectively learn the different data distribution characteristics between the two domains using only the data from the source domain and the unlabeled target domain data. The representative visual segmentation examples were presented in Fig. \ref{RIGA+_visual}. In contrast to existing UDA methodologies, our method obtained more reliable predictions with a reduced incidence of false negatives. In addition, as can be seen from Fig. \ref{RIGA+_visual}, in the results of the OD segmentation, our method outperformed the others in terms of boundary delineation and fine\nobreakdash-grained detail handling.
\par Additionally, we performed ablation experiments on the RIGA+ dataset to validate the importance of each module shown in Table \ref{tab:performance_comparison}. Specifically, ``FullNet" outperformed ``IntraOnly", ``CrossOnly (Uni)", ``CrossOnly" and ``CrossUni" on the average Dice metric by 2.76\%, 1\%, 0.44\% and 0.29\%, respectively.

\section{Conclusion}\label{conclusion}
In this paper, we propose the DDS\nobreakdash-UDA for unsupervised domain adaptation to jointly segment optic disc and optic cup, which is based on integrating bi\nobreakdash-directional cross\nobreakdash-domain consistency and intra\nobreakdash-domain pseudo label learning. The proposed bi\nobreakdash-directional cross\nobreakdash-domain consistency utilizes dynamic feature\nobreakdash-level masking to progressively suppress domain\nobreakdash-specific noise while preserving structural semantics through coarse\nobreakdash-to\nobreakdash-fine feature exchange. On the other hand, the frequency\nobreakdash-driven pseudo label learning strategy enriches the intra\nobreakdash-domain diversity through spectral amplitude mixing to ensure high\nobreakdash-fidelity supervision for robust feature alignment. By jointly optimizing these components, DDS\nobreakdash-UDA can efficiently extract domain\nobreakdash-invariant features and mitigate the domain shift problem for multi\nobreakdash-domain datasets. The comprehensive experiments showed that our framework achieves superior segmentation performance and generalization ability over several current approaches. In future work, we will explore extending DDS\nobreakdash-UDA with stronger backbone architectures and complementary priors to further improve robustness and generalization across challenging multi-centre settings.

\section*{Data availability}
All the data used by this study are publicly available from the references.

\section*{Acknowledgments}
This work was supported in part by the National Natural Science Foundation of China under Grants 62202442 and 12261059, in part by the Natural Science Foundation of Jiangxi Province under Grant 20224BAB211001, in part by the NIH under Grants R01GM109068, R01MH104680, R01MH107354, R01EB006841, P20GM103472, and U19AG055373, and in part by the NSF under Grant 1539067.

% \bibliographystyle{elsarticle-harv}
% \biboptions{authoryear}
% \bibliography{reference}

\end{document}